\documentclass[10pt]{article}

\usepackage[top=0.75in, bottom=0.75in, left=1in, right=1in]{geometry}

\usepackage{setspace}

\usepackage{amsmath,amsthm,amssymb}
\usepackage{atbegshi,picture}
\usepackage{url}
\usepackage[pdftex]{graphicx} 
\DeclareGraphicsExtensions{.pdf,.png,.jpg}
\usepackage{sidecap} 
\usepackage{tikz} 
\usepackage[font=small,labelfont=bf]{caption}
\usepackage{listings}
\usepackage{indentfirst}
\usepackage{color}
\usepackage[draft]{minted}
\usepackage{pgfplots}

\makeatletter
\def\PYG@reset{\let\PYG@it=\relax \let\PYG@bf=\relax%
    \let\PYG@ul=\relax \let\PYG@tc=\relax%
    \let\PYG@bc=\relax \let\PYG@ff=\relax}
\def\PYG@tok#1{\csname PYG@tok@#1\endcsname}
\def\PYG@toks#1+{\ifx\relax#1\empty\else%
    \PYG@tok{#1}\expandafter\PYG@toks\fi}
\def\PYG@do#1{\PYG@bc{\PYG@tc{\PYG@ul{%
    \PYG@it{\PYG@bf{\PYG@ff{#1}}}}}}}
\def\PYG#1#2{\PYG@reset\PYG@toks#1+\relax+\PYG@do{#2}}

\expandafter\def\csname PYG@tok@w\endcsname{\def\PYG@tc##1{\textcolor[rgb]{0.73,0.73,0.73}{##1}}}
\expandafter\def\csname PYG@tok@c\endcsname{\let\PYG@it=\textit\def\PYG@tc##1{\textcolor[rgb]{0.25,0.50,0.50}{##1}}}
\expandafter\def\csname PYG@tok@cp\endcsname{\def\PYG@tc##1{\textcolor[rgb]{0.74,0.48,0.00}{##1}}}
\expandafter\def\csname PYG@tok@k\endcsname{\let\PYG@bf=\textbf\def\PYG@tc##1{\textcolor[rgb]{0.00,0.50,0.00}{##1}}}
\expandafter\def\csname PYG@tok@kp\endcsname{\def\PYG@tc##1{\textcolor[rgb]{0.00,0.50,0.00}{##1}}}
\expandafter\def\csname PYG@tok@kt\endcsname{\def\PYG@tc##1{\textcolor[rgb]{0.69,0.00,0.25}{##1}}}
\expandafter\def\csname PYG@tok@o\endcsname{\def\PYG@tc##1{\textcolor[rgb]{0.40,0.40,0.40}{##1}}}
\expandafter\def\csname PYG@tok@ow\endcsname{\let\PYG@bf=\textbf\def\PYG@tc##1{\textcolor[rgb]{0.67,0.13,1.00}{##1}}}
\expandafter\def\csname PYG@tok@nb\endcsname{\def\PYG@tc##1{\textcolor[rgb]{0.00,0.50,0.00}{##1}}}
\expandafter\def\csname PYG@tok@nf\endcsname{\def\PYG@tc##1{\textcolor[rgb]{0.00,0.00,1.00}{##1}}}
\expandafter\def\csname PYG@tok@nc\endcsname{\let\PYG@bf=\textbf\def\PYG@tc##1{\textcolor[rgb]{0.00,0.00,1.00}{##1}}}
\expandafter\def\csname PYG@tok@nn\endcsname{\let\PYG@bf=\textbf\def\PYG@tc##1{\textcolor[rgb]{0.00,0.00,1.00}{##1}}}
\expandafter\def\csname PYG@tok@ne\endcsname{\let\PYG@bf=\textbf\def\PYG@tc##1{\textcolor[rgb]{0.82,0.25,0.23}{##1}}}
\expandafter\def\csname PYG@tok@nv\endcsname{\def\PYG@tc##1{\textcolor[rgb]{0.10,0.09,0.49}{##1}}}
\expandafter\def\csname PYG@tok@no\endcsname{\def\PYG@tc##1{\textcolor[rgb]{0.53,0.00,0.00}{##1}}}
\expandafter\def\csname PYG@tok@nl\endcsname{\def\PYG@tc##1{\textcolor[rgb]{0.63,0.63,0.00}{##1}}}
\expandafter\def\csname PYG@tok@ni\endcsname{\let\PYG@bf=\textbf\def\PYG@tc##1{\textcolor[rgb]{0.60,0.60,0.60}{##1}}}
\expandafter\def\csname PYG@tok@na\endcsname{\def\PYG@tc##1{\textcolor[rgb]{0.49,0.56,0.16}{##1}}}
\expandafter\def\csname PYG@tok@nt\endcsname{\let\PYG@bf=\textbf\def\PYG@tc##1{\textcolor[rgb]{0.00,0.50,0.00}{##1}}}
\expandafter\def\csname PYG@tok@nd\endcsname{\def\PYG@tc##1{\textcolor[rgb]{0.67,0.13,1.00}{##1}}}
\expandafter\def\csname PYG@tok@s\endcsname{\def\PYG@tc##1{\textcolor[rgb]{0.73,0.13,0.13}{##1}}}
\expandafter\def\csname PYG@tok@sd\endcsname{\let\PYG@it=\textit\def\PYG@tc##1{\textcolor[rgb]{0.73,0.13,0.13}{##1}}}
\expandafter\def\csname PYG@tok@si\endcsname{\let\PYG@bf=\textbf\def\PYG@tc##1{\textcolor[rgb]{0.73,0.40,0.53}{##1}}}
\expandafter\def\csname PYG@tok@se\endcsname{\let\PYG@bf=\textbf\def\PYG@tc##1{\textcolor[rgb]{0.73,0.40,0.13}{##1}}}
\expandafter\def\csname PYG@tok@sr\endcsname{\def\PYG@tc##1{\textcolor[rgb]{0.73,0.40,0.53}{##1}}}
\expandafter\def\csname PYG@tok@ss\endcsname{\def\PYG@tc##1{\textcolor[rgb]{0.10,0.09,0.49}{##1}}}
\expandafter\def\csname PYG@tok@sx\endcsname{\def\PYG@tc##1{\textcolor[rgb]{0.00,0.50,0.00}{##1}}}
\expandafter\def\csname PYG@tok@m\endcsname{\def\PYG@tc##1{\textcolor[rgb]{0.40,0.40,0.40}{##1}}}
\expandafter\def\csname PYG@tok@gh\endcsname{\let\PYG@bf=\textbf\def\PYG@tc##1{\textcolor[rgb]{0.00,0.00,0.50}{##1}}}
\expandafter\def\csname PYG@tok@gu\endcsname{\let\PYG@bf=\textbf\def\PYG@tc##1{\textcolor[rgb]{0.50,0.00,0.50}{##1}}}
\expandafter\def\csname PYG@tok@gd\endcsname{\def\PYG@tc##1{\textcolor[rgb]{0.63,0.00,0.00}{##1}}}
\expandafter\def\csname PYG@tok@gi\endcsname{\def\PYG@tc##1{\textcolor[rgb]{0.00,0.63,0.00}{##1}}}
\expandafter\def\csname PYG@tok@gr\endcsname{\def\PYG@tc##1{\textcolor[rgb]{1.00,0.00,0.00}{##1}}}
\expandafter\def\csname PYG@tok@ge\endcsname{\let\PYG@it=\textit}
\expandafter\def\csname PYG@tok@gs\endcsname{\let\PYG@bf=\textbf}
\expandafter\def\csname PYG@tok@gp\endcsname{\let\PYG@bf=\textbf\def\PYG@tc##1{\textcolor[rgb]{0.00,0.00,0.50}{##1}}}
\expandafter\def\csname PYG@tok@go\endcsname{\def\PYG@tc##1{\textcolor[rgb]{0.53,0.53,0.53}{##1}}}
\expandafter\def\csname PYG@tok@gt\endcsname{\def\PYG@tc##1{\textcolor[rgb]{0.00,0.27,0.87}{##1}}}
\expandafter\def\csname PYG@tok@err\endcsname{\def\PYG@bc##1{\setlength{\fboxsep}{0pt}\fcolorbox[rgb]{1.00,0.00,0.00}{1,1,1}{\strut ##1}}}
\expandafter\def\csname PYG@tok@kc\endcsname{\let\PYG@bf=\textbf\def\PYG@tc##1{\textcolor[rgb]{0.00,0.50,0.00}{##1}}}
\expandafter\def\csname PYG@tok@kd\endcsname{\let\PYG@bf=\textbf\def\PYG@tc##1{\textcolor[rgb]{0.00,0.50,0.00}{##1}}}
\expandafter\def\csname PYG@tok@kn\endcsname{\let\PYG@bf=\textbf\def\PYG@tc##1{\textcolor[rgb]{0.00,0.50,0.00}{##1}}}
\expandafter\def\csname PYG@tok@kr\endcsname{\let\PYG@bf=\textbf\def\PYG@tc##1{\textcolor[rgb]{0.00,0.50,0.00}{##1}}}
\expandafter\def\csname PYG@tok@bp\endcsname{\def\PYG@tc##1{\textcolor[rgb]{0.00,0.50,0.00}{##1}}}
\expandafter\def\csname PYG@tok@fm\endcsname{\def\PYG@tc##1{\textcolor[rgb]{0.00,0.00,1.00}{##1}}}
\expandafter\def\csname PYG@tok@vc\endcsname{\def\PYG@tc##1{\textcolor[rgb]{0.10,0.09,0.49}{##1}}}
\expandafter\def\csname PYG@tok@vg\endcsname{\def\PYG@tc##1{\textcolor[rgb]{0.10,0.09,0.49}{##1}}}
\expandafter\def\csname PYG@tok@vi\endcsname{\def\PYG@tc##1{\textcolor[rgb]{0.10,0.09,0.49}{##1}}}
\expandafter\def\csname PYG@tok@vm\endcsname{\def\PYG@tc##1{\textcolor[rgb]{0.10,0.09,0.49}{##1}}}
\expandafter\def\csname PYG@tok@sa\endcsname{\def\PYG@tc##1{\textcolor[rgb]{0.73,0.13,0.13}{##1}}}
\expandafter\def\csname PYG@tok@sb\endcsname{\def\PYG@tc##1{\textcolor[rgb]{0.73,0.13,0.13}{##1}}}
\expandafter\def\csname PYG@tok@sc\endcsname{\def\PYG@tc##1{\textcolor[rgb]{0.73,0.13,0.13}{##1}}}
\expandafter\def\csname PYG@tok@dl\endcsname{\def\PYG@tc##1{\textcolor[rgb]{0.73,0.13,0.13}{##1}}}
\expandafter\def\csname PYG@tok@s2\endcsname{\def\PYG@tc##1{\textcolor[rgb]{0.73,0.13,0.13}{##1}}}
\expandafter\def\csname PYG@tok@sh\endcsname{\def\PYG@tc##1{\textcolor[rgb]{0.73,0.13,0.13}{##1}}}
\expandafter\def\csname PYG@tok@s1\endcsname{\def\PYG@tc##1{\textcolor[rgb]{0.73,0.13,0.13}{##1}}}
\expandafter\def\csname PYG@tok@mb\endcsname{\def\PYG@tc##1{\textcolor[rgb]{0.40,0.40,0.40}{##1}}}
\expandafter\def\csname PYG@tok@mf\endcsname{\def\PYG@tc##1{\textcolor[rgb]{0.40,0.40,0.40}{##1}}}
\expandafter\def\csname PYG@tok@mh\endcsname{\def\PYG@tc##1{\textcolor[rgb]{0.40,0.40,0.40}{##1}}}
\expandafter\def\csname PYG@tok@mi\endcsname{\def\PYG@tc##1{\textcolor[rgb]{0.40,0.40,0.40}{##1}}}
\expandafter\def\csname PYG@tok@il\endcsname{\def\PYG@tc##1{\textcolor[rgb]{0.40,0.40,0.40}{##1}}}
\expandafter\def\csname PYG@tok@mo\endcsname{\def\PYG@tc##1{\textcolor[rgb]{0.40,0.40,0.40}{##1}}}
\expandafter\def\csname PYG@tok@ch\endcsname{\let\PYG@it=\textit\def\PYG@tc##1{\textcolor[rgb]{0.25,0.50,0.50}{##1}}}
\expandafter\def\csname PYG@tok@cm\endcsname{\let\PYG@it=\textit\def\PYG@tc##1{\textcolor[rgb]{0.25,0.50,0.50}{##1}}}
\expandafter\def\csname PYG@tok@cpf\endcsname{\let\PYG@it=\textit\def\PYG@tc##1{\textcolor[rgb]{0.25,0.50,0.50}{##1}}}
\expandafter\def\csname PYG@tok@c1\endcsname{\let\PYG@it=\textit\def\PYG@tc##1{\textcolor[rgb]{0.25,0.50,0.50}{##1}}}
\expandafter\def\csname PYG@tok@cs\endcsname{\let\PYG@it=\textit\def\PYG@tc##1{\textcolor[rgb]{0.25,0.50,0.50}{##1}}}

\def\PYGZus{\char`\_}
\def\PYGZob{\char`\{}
\def\PYGZcb{\char`\}}

\def\PYGZlt{\char`\<}
\def\PYGZgt{\char`\>}

\def\PYGZhy{\char`\-}

\makeatother

\makeatletter
\def\PYGdefault@reset{\let\PYGdefault@it=\relax \let\PYGdefault@bf=\relax%
    \let\PYGdefault@ul=\relax \let\PYGdefault@tc=\relax%
    \let\PYGdefault@bc=\relax \let\PYGdefault@ff=\relax}
\def\PYGdefault@tok#1{\csname PYGdefault@tok@#1\endcsname}
\def\PYGdefault@toks#1+{\ifx\relax#1\empty\else%
    \PYGdefault@tok{#1}\expandafter\PYGdefault@toks\fi}
\def\PYGdefault@do#1{\PYGdefault@bc{\PYGdefault@tc{\PYGdefault@ul{%
    \PYGdefault@it{\PYGdefault@bf{\PYGdefault@ff{#1}}}}}}}
\def\PYGdefault#1#2{\PYGdefault@reset\PYGdefault@toks#1+\relax+\PYGdefault@do{#2}}

\expandafter\def\csname PYGdefault@tok@w\endcsname{\def\PYGdefault@tc##1{\textcolor[rgb]{0.73,0.73,0.73}{##1}}}
\expandafter\def\csname PYGdefault@tok@c\endcsname{\let\PYGdefault@it=\textit\def\PYGdefault@tc##1{\textcolor[rgb]{0.25,0.50,0.50}{##1}}}
\expandafter\def\csname PYGdefault@tok@cp\endcsname{\def\PYGdefault@tc##1{\textcolor[rgb]{0.74,0.48,0.00}{##1}}}
\expandafter\def\csname PYGdefault@tok@k\endcsname{\let\PYGdefault@bf=\textbf\def\PYGdefault@tc##1{\textcolor[rgb]{0.00,0.50,0.00}{##1}}}
\expandafter\def\csname PYGdefault@tok@kp\endcsname{\def\PYGdefault@tc##1{\textcolor[rgb]{0.00,0.50,0.00}{##1}}}
\expandafter\def\csname PYGdefault@tok@kt\endcsname{\def\PYGdefault@tc##1{\textcolor[rgb]{0.69,0.00,0.25}{##1}}}
\expandafter\def\csname PYGdefault@tok@o\endcsname{\def\PYGdefault@tc##1{\textcolor[rgb]{0.40,0.40,0.40}{##1}}}
\expandafter\def\csname PYGdefault@tok@ow\endcsname{\let\PYGdefault@bf=\textbf\def\PYGdefault@tc##1{\textcolor[rgb]{0.67,0.13,1.00}{##1}}}
\expandafter\def\csname PYGdefault@tok@nb\endcsname{\def\PYGdefault@tc##1{\textcolor[rgb]{0.00,0.50,0.00}{##1}}}
\expandafter\def\csname PYGdefault@tok@nf\endcsname{\def\PYGdefault@tc##1{\textcolor[rgb]{0.00,0.00,1.00}{##1}}}
\expandafter\def\csname PYGdefault@tok@nc\endcsname{\let\PYGdefault@bf=\textbf\def\PYGdefault@tc##1{\textcolor[rgb]{0.00,0.00,1.00}{##1}}}
\expandafter\def\csname PYGdefault@tok@nn\endcsname{\let\PYGdefault@bf=\textbf\def\PYGdefault@tc##1{\textcolor[rgb]{0.00,0.00,1.00}{##1}}}
\expandafter\def\csname PYGdefault@tok@ne\endcsname{\let\PYGdefault@bf=\textbf\def\PYGdefault@tc##1{\textcolor[rgb]{0.82,0.25,0.23}{##1}}}
\expandafter\def\csname PYGdefault@tok@nv\endcsname{\def\PYGdefault@tc##1{\textcolor[rgb]{0.10,0.09,0.49}{##1}}}
\expandafter\def\csname PYGdefault@tok@no\endcsname{\def\PYGdefault@tc##1{\textcolor[rgb]{0.53,0.00,0.00}{##1}}}
\expandafter\def\csname PYGdefault@tok@nl\endcsname{\def\PYGdefault@tc##1{\textcolor[rgb]{0.63,0.63,0.00}{##1}}}
\expandafter\def\csname PYGdefault@tok@ni\endcsname{\let\PYGdefault@bf=\textbf\def\PYGdefault@tc##1{\textcolor[rgb]{0.60,0.60,0.60}{##1}}}
\expandafter\def\csname PYGdefault@tok@na\endcsname{\def\PYGdefault@tc##1{\textcolor[rgb]{0.49,0.56,0.16}{##1}}}
\expandafter\def\csname PYGdefault@tok@nt\endcsname{\let\PYGdefault@bf=\textbf\def\PYGdefault@tc##1{\textcolor[rgb]{0.00,0.50,0.00}{##1}}}
\expandafter\def\csname PYGdefault@tok@nd\endcsname{\def\PYGdefault@tc##1{\textcolor[rgb]{0.67,0.13,1.00}{##1}}}
\expandafter\def\csname PYGdefault@tok@s\endcsname{\def\PYGdefault@tc##1{\textcolor[rgb]{0.73,0.13,0.13}{##1}}}
\expandafter\def\csname PYGdefault@tok@sd\endcsname{\let\PYGdefault@it=\textit\def\PYGdefault@tc##1{\textcolor[rgb]{0.73,0.13,0.13}{##1}}}
\expandafter\def\csname PYGdefault@tok@si\endcsname{\let\PYGdefault@bf=\textbf\def\PYGdefault@tc##1{\textcolor[rgb]{0.73,0.40,0.53}{##1}}}
\expandafter\def\csname PYGdefault@tok@se\endcsname{\let\PYGdefault@bf=\textbf\def\PYGdefault@tc##1{\textcolor[rgb]{0.73,0.40,0.13}{##1}}}
\expandafter\def\csname PYGdefault@tok@sr\endcsname{\def\PYGdefault@tc##1{\textcolor[rgb]{0.73,0.40,0.53}{##1}}}
\expandafter\def\csname PYGdefault@tok@ss\endcsname{\def\PYGdefault@tc##1{\textcolor[rgb]{0.10,0.09,0.49}{##1}}}
\expandafter\def\csname PYGdefault@tok@sx\endcsname{\def\PYGdefault@tc##1{\textcolor[rgb]{0.00,0.50,0.00}{##1}}}
\expandafter\def\csname PYGdefault@tok@m\endcsname{\def\PYGdefault@tc##1{\textcolor[rgb]{0.40,0.40,0.40}{##1}}}
\expandafter\def\csname PYGdefault@tok@gh\endcsname{\let\PYGdefault@bf=\textbf\def\PYGdefault@tc##1{\textcolor[rgb]{0.00,0.00,0.50}{##1}}}
\expandafter\def\csname PYGdefault@tok@gu\endcsname{\let\PYGdefault@bf=\textbf\def\PYGdefault@tc##1{\textcolor[rgb]{0.50,0.00,0.50}{##1}}}
\expandafter\def\csname PYGdefault@tok@gd\endcsname{\def\PYGdefault@tc##1{\textcolor[rgb]{0.63,0.00,0.00}{##1}}}
\expandafter\def\csname PYGdefault@tok@gi\endcsname{\def\PYGdefault@tc##1{\textcolor[rgb]{0.00,0.63,0.00}{##1}}}
\expandafter\def\csname PYGdefault@tok@gr\endcsname{\def\PYGdefault@tc##1{\textcolor[rgb]{1.00,0.00,0.00}{##1}}}
\expandafter\def\csname PYGdefault@tok@ge\endcsname{\let\PYGdefault@it=\textit}
\expandafter\def\csname PYGdefault@tok@gs\endcsname{\let\PYGdefault@bf=\textbf}
\expandafter\def\csname PYGdefault@tok@gp\endcsname{\let\PYGdefault@bf=\textbf\def\PYGdefault@tc##1{\textcolor[rgb]{0.00,0.00,0.50}{##1}}}
\expandafter\def\csname PYGdefault@tok@go\endcsname{\def\PYGdefault@tc##1{\textcolor[rgb]{0.53,0.53,0.53}{##1}}}
\expandafter\def\csname PYGdefault@tok@gt\endcsname{\def\PYGdefault@tc##1{\textcolor[rgb]{0.00,0.27,0.87}{##1}}}
\expandafter\def\csname PYGdefault@tok@err\endcsname{\def\PYGdefault@bc##1{\setlength{\fboxsep}{0pt}\fcolorbox[rgb]{1.00,0.00,0.00}{1,1,1}{\strut ##1}}}
\expandafter\def\csname PYGdefault@tok@kc\endcsname{\let\PYGdefault@bf=\textbf\def\PYGdefault@tc##1{\textcolor[rgb]{0.00,0.50,0.00}{##1}}}
\expandafter\def\csname PYGdefault@tok@kd\endcsname{\let\PYGdefault@bf=\textbf\def\PYGdefault@tc##1{\textcolor[rgb]{0.00,0.50,0.00}{##1}}}
\expandafter\def\csname PYGdefault@tok@kn\endcsname{\let\PYGdefault@bf=\textbf\def\PYGdefault@tc##1{\textcolor[rgb]{0.00,0.50,0.00}{##1}}}
\expandafter\def\csname PYGdefault@tok@kr\endcsname{\let\PYGdefault@bf=\textbf\def\PYGdefault@tc##1{\textcolor[rgb]{0.00,0.50,0.00}{##1}}}
\expandafter\def\csname PYGdefault@tok@bp\endcsname{\def\PYGdefault@tc##1{\textcolor[rgb]{0.00,0.50,0.00}{##1}}}
\expandafter\def\csname PYGdefault@tok@fm\endcsname{\def\PYGdefault@tc##1{\textcolor[rgb]{0.00,0.00,1.00}{##1}}}
\expandafter\def\csname PYGdefault@tok@vc\endcsname{\def\PYGdefault@tc##1{\textcolor[rgb]{0.10,0.09,0.49}{##1}}}
\expandafter\def\csname PYGdefault@tok@vg\endcsname{\def\PYGdefault@tc##1{\textcolor[rgb]{0.10,0.09,0.49}{##1}}}
\expandafter\def\csname PYGdefault@tok@vi\endcsname{\def\PYGdefault@tc##1{\textcolor[rgb]{0.10,0.09,0.49}{##1}}}
\expandafter\def\csname PYGdefault@tok@vm\endcsname{\def\PYGdefault@tc##1{\textcolor[rgb]{0.10,0.09,0.49}{##1}}}
\expandafter\def\csname PYGdefault@tok@sa\endcsname{\def\PYGdefault@tc##1{\textcolor[rgb]{0.73,0.13,0.13}{##1}}}
\expandafter\def\csname PYGdefault@tok@sb\endcsname{\def\PYGdefault@tc##1{\textcolor[rgb]{0.73,0.13,0.13}{##1}}}
\expandafter\def\csname PYGdefault@tok@sc\endcsname{\def\PYGdefault@tc##1{\textcolor[rgb]{0.73,0.13,0.13}{##1}}}
\expandafter\def\csname PYGdefault@tok@dl\endcsname{\def\PYGdefault@tc##1{\textcolor[rgb]{0.73,0.13,0.13}{##1}}}
\expandafter\def\csname PYGdefault@tok@s2\endcsname{\def\PYGdefault@tc##1{\textcolor[rgb]{0.73,0.13,0.13}{##1}}}
\expandafter\def\csname PYGdefault@tok@sh\endcsname{\def\PYGdefault@tc##1{\textcolor[rgb]{0.73,0.13,0.13}{##1}}}
\expandafter\def\csname PYGdefault@tok@s1\endcsname{\def\PYGdefault@tc##1{\textcolor[rgb]{0.73,0.13,0.13}{##1}}}
\expandafter\def\csname PYGdefault@tok@mb\endcsname{\def\PYGdefault@tc##1{\textcolor[rgb]{0.40,0.40,0.40}{##1}}}
\expandafter\def\csname PYGdefault@tok@mf\endcsname{\def\PYGdefault@tc##1{\textcolor[rgb]{0.40,0.40,0.40}{##1}}}
\expandafter\def\csname PYGdefault@tok@mh\endcsname{\def\PYGdefault@tc##1{\textcolor[rgb]{0.40,0.40,0.40}{##1}}}
\expandafter\def\csname PYGdefault@tok@mi\endcsname{\def\PYGdefault@tc##1{\textcolor[rgb]{0.40,0.40,0.40}{##1}}}
\expandafter\def\csname PYGdefault@tok@il\endcsname{\def\PYGdefault@tc##1{\textcolor[rgb]{0.40,0.40,0.40}{##1}}}
\expandafter\def\csname PYGdefault@tok@mo\endcsname{\def\PYGdefault@tc##1{\textcolor[rgb]{0.40,0.40,0.40}{##1}}}
\expandafter\def\csname PYGdefault@tok@ch\endcsname{\let\PYGdefault@it=\textit\def\PYGdefault@tc##1{\textcolor[rgb]{0.25,0.50,0.50}{##1}}}
\expandafter\def\csname PYGdefault@tok@cm\endcsname{\let\PYGdefault@it=\textit\def\PYGdefault@tc##1{\textcolor[rgb]{0.25,0.50,0.50}{##1}}}
\expandafter\def\csname PYGdefault@tok@cpf\endcsname{\let\PYGdefault@it=\textit\def\PYGdefault@tc##1{\textcolor[rgb]{0.25,0.50,0.50}{##1}}}
\expandafter\def\csname PYGdefault@tok@c1\endcsname{\let\PYGdefault@it=\textit\def\PYGdefault@tc##1{\textcolor[rgb]{0.25,0.50,0.50}{##1}}}
\expandafter\def\csname PYGdefault@tok@cs\endcsname{\let\PYGdefault@it=\textit\def\PYGdefault@tc##1{\textcolor[rgb]{0.25,0.50,0.50}{##1}}}

\makeatother

\usepackage{pbox}
\usepackage{fancyvrb}
\usepackage[titletoc]{appendix}

\begin{document}

\begin{titlepage}
\title{An Analysis of Parallelized Motion Masking Using Dual-Mode Single Gaussian Models}
\author{Peter Henderson\footnote{peter.henderson@mail.mcgill.ca} \hspace{1cm} Matthew Vertescher}
\date{}
\maketitle
\thispagestyle{empty} 
\end{titlepage}

\abstract{Motion detection in video is important for a number of applications and fields. In video surveillance, motion detection is an essential accompaniment to activity recognition for early warning systems. Robotics also has much to gain from motion detection and segmentation, particularly in high speed motion tracking for tactile systems. There are a myriad of techniques for detecting and masking motion in an image. Successful systems have used Gaussian Models to discern background from foreground in an image (motion from static imagery). However, particularly in the case of a moving camera or frame of reference, it is necessary to compensate for the motion of the camera when attempting to discern objects moving in the foreground. For example, it is possible to estimate motion of the camera through optical flow methods or temporal differencing and then compensate for this motion in a background subtraction model. We selection a method by Yi et al. using Dual-Mode Single Gaussian Models which does just this. We implement the technique in Intel's Thread Building Blocks (TBB) and NVIDIA's CUDA libraries. We then compare parallelization improvements with a theoretical analysis of speedups based on the characteristics of our selected model and attributes of both TBB and CUDA. We make our implementation available to the public.}

\newpage

\tableofcontents

\newpage


\section{Introduction}
Motion detection in video is important for a number of applications and fields. In video surveillance, motion detection is an essential accompaniment to activity recognition for early warning systems~\cite{huang2011advanced, haritaoglu2000w, yu2009real}. Robotics also has much to gain from motion detection and segmentation, particularly in high speed motion tracking for tactile systems~\cite{senoo2014high}. Others can clearly benefit from such technologies as well. However, motion detection is a hard problem, particularly with a moving camera reference frame, frequent changes in lighting or dynamic backgrounds. As such, many attempts at tackling this problem have been made. Modern research into motion detection and segmentation systems generally falls into 3 categories~\cite{hu2004survey}: background subtraction, temporal differencing, and optical flow.

Background subtraction involves developing a model in a series of images which differentiates the foreground (i.e. motion) from its background. A background model is typically learned on a per-pixel basis. A large portion of background modeling techniques involve keeping a Gaussian Model (GM), or number of candidate GMs, of the background for each pixel. Wren et al.~\cite{wren1997pfinder} first used Single Gaussian Models (SGM) (with one GM per pixel) for motion detection. The Gaussian model is updated incrementally per some learning rate depending on whether or not a pixel intensity falls within the model's variance. While this technique provided acceptable results for certain cases, large objects moving through the image or changes in lighting proves to contaminate the model significantly. Further expansion on this involves keeping several candidate Gaussian models indicating possible backgrounds, and a probabilistic mixture is used to differentiate the foreground from the background~\cite{stauffer1999adaptive}. This typically yields better results in most cases, but can still run into the same problems as SGMs.

In temporal differencing, the absolute difference of pixel intensities across two or more frames is averaged and a threshold value is used to determine when motion occurs. Component analysis methods are used to cluster differences into regions of motion~\cite{lipton1998moving, yu2009real}.

Optical flow methods are widely varied, but there are several methods used often as a basis for many complex systems. Two often discussed algorithms are Lucas-Kanade~\cite{lucas1981iterative} and Horn-Schunk~\cite{horn1981determining}. Lucas-Kanade find the vector transformation of pixels by using least-squares methods to determine the flow for a neighborhood of pixels, while Horn-Schunk used the smoothness constraint in the aperture problem to find optical flow in an image.

Many newer motion detection systems use an ensemble of these methods to accomplish the complex task of motion detection. Particularly in the case of a moving camera or frame of reference, it is necessary to compensate for the motion of the camera when attempting to discern objects moving in the foreground. For example, it is possible to estimate motion of the camera through optical flow methods or temporal differencing and then compensate for this motion in a background subtraction model~\cite{yi2013detection, kim2013detection}.

We select a method by Yi et al.~\cite{yi2013detection} to detect motion in a series of images (or video) using one of these ensemble methods (though the focus is on background subtraction methods). We use OpenCV for some basic tools (like image parsing and display) as well as some of the motion compensation and preprocessing. We parallelize the motion detection algorithm using Intel's Thread Building Blocks (TBB) and NVIDIA's Compute Unified Device Architecture (CUDA) libraries.

Intel's TBB is a library which provides abstractions for interaction with operating system provided threading libraries (i.e. POSIX)~\cite{tbb1, tbb2}. Rather than dealing with thread synchronization and handling manually, TBB provides an abstraction which allows for the creation of tasks which are allocated to different machine cores by the library run-time engine. TBB also handles CPU-cache transactions to make processing more efficient. We choose TBB because of the abstractions available to simplify parallelization as well as the optimizations of task scheduling and cache management which the library provides. We also considered OpenMP~\cite{omp}, but ran into compilation issues relating to the Mac OS X gcc compiler's interaction with OpenCV. OpenMP is not supported by older versions of the gcc tool on Mac. However, newer versions of the tool are not compatible with OpenCV on Mac. We then reverted to using TBB as it seemed like a sensible and powerful alternative.

NVIDIA's CUDA is a platform which gives developers access to low level GPU instruction sets~\cite{aboutCUDA}. References to CUDA often refer to the NVIDIA CUDA C/C++ language and libraries which provide high level control over the GPU. It is this that we use for our project. While this can vary between graphics cards, generally the paradigm for CUDA GPUs is as such (based on the Kepler architecture~\cite{keplerWhitePaper}, but applies similarly to others like Fermi~\cite{fermiWhitePaper}). The GPU contains several multiprocessors (MPs) each with many CUDA cores (processing units). Processing functions are divided into blocks of threads. Each block is assigned to one MP. Once a block is assigned, another MP cannot access that block or its shared memory. The block of threads is further assigned to CUDA cores by a warp scheduler. A warp is a group of threads which will be scheduled onto CUDA cores in parallel, typically in blocks of 32, but this depends on the warp size of the card. The threads can either fetch from global or shared memory, but shared memory is typically shared only within a block on an MP.

\section{Background}

\subsection{Preprocessing}

To improve the accuracy of the motion detection, as in Yi et al., we first pre-process the incoming frame. Two filters are used for this task: a Gaussian filter and a median filter. A Gaussian filter creates a matrix of size $N \times N$ with a Gaussian distribution of weights across the matrix. The Gaussian distribution function for a one dimensional vector can be seen in Equation~\ref{eq:gauss} where $\sigma$ is the standard deviation, $\mu$ is the mean value and $x$ is the distance parameter desired.
\begin{equation}
\frac{1}{{\sigma \sqrt {2\pi } }}e^{{{ - \left( {x - \mu } \right)^2 } \mathord{\left/ {\vphantom {{ - \left( {x - \mu } \right)^2 } {2\sigma ^2 }}} \right. \kern-\nulldelimiterspace} {2\sigma ^2 }}}
\label{eq:gauss}
\end{equation}

The image can then be convolved with this normal distribution of weights to set a pixel's value to the summed normal distribution of its neighbors for some given size and sigma value.

In a median filter, a pixel is set to the median value of its neighbors. This preprocessing is done to smooth the edges of motion detection and ensure that fewer per-pixel lighting anomalies are detected.

\subsection{Single Gaussian Model}

In SGM background subtraction methods, the mean and variance of every pixel in a series of images is kept. They are updated based on a specified learning rate to develop a model for the background of an image. In their method, Yi et al. use a variable learning rate based on the age of the model to place more emphasis on earlier frames. However, the age is capped at 30 to prevent excessive emphasis and ghosting effects\footnote{Ghosting is referred to contamination of the background model such that motion is always observed there because a previously stationary object has moved from the background, leaving a ghosting effect.}. On each iteration which affects the background model, the mean, variance and age are updated as follows. We take the equations here directly from Yi et al.'s paper. However, it is important to note that while Yi et al. use an average of pixel intensities in a grid of size size $m\times m$, we calculate these on a per pixel basis, and hence, for each grid the size of our Grid is 1:

\begin{equation}
|G_i| = 1
\end{equation}

The mean of the SGM is updated according to:

\begin{equation}
\mu_{i}^{(t)} = \frac{\tilde{\alpha}_{i}^{(t-1)} }{ \tilde{\alpha}_{i}^{(t-1)} + 1 } 	\tilde{\mu}_{i}^{(t-1)} + \frac{1}{\tilde{\alpha}_{i}^{(t-1)} + 1} M_{i}^{t}
\end{equation}

Here, the mean of the background model $\mu_{i}^{(t)}$ is updated based on the previous mean and the current pixel intensity. While Yi et al. use $M_{i}^{t}$ to be defined as follows:

\begin{equation}
M_{i}^{(t)} = \frac{1}{|G_i|} \sum\limits_{j \in G_i}^{} I_{j}^{(t)} \\
\end{equation}

We use a per-pixel model, and thus $M_{i}^{t}$ is equivalent to $I_{j}^{(t)}$. Further, the variance is updated according to:

\begin{equation}
\sigma_{i}^{(t)} = \frac{\tilde{\alpha}_{i}^{(t-1)} }{ \tilde{\alpha}_{i}^{(t-1)} + 1 } 	\tilde{\sigma}_{i}^{(t-1)} + \frac{1}{\tilde{\alpha}_{i}^{(t-1)} + 1} V_{i}^{t} \\
\end{equation}

Here, the variance $\sigma_{i}^{(t)}$ is updated based on the age, the previous variance, and the absolute difference of the mean and the current pixel intensities. Again, though Yi et al., define:

\begin{equation}
V_{i}^{(t)} = \max_{j \in G_i} \left( \mu_{i}^{(t)} - I_{j}^{(t)} \right)^2
\end{equation}

We use a per-pixel model, and as such this just becomes $\left( \mu_{i}^{(t)} - I_{j}^{(t)} \right)^2$. Lastly, the age of the model is updated according to:

\begin{equation}
\alpha_{i}^{(t)} = \tilde{\alpha}_{i}^{(t-1)} + 1 \\
\end{equation}

In all of the above, $\tilde{\alpha}, \tilde{\mu}, \tilde{\sigma}$ represent the previous values after motion compensation, as will be discussed later. While an SGM may be acceptable for a static background and relatively simple background subtraction tasks, large objects moving through the image sequences can contaminate the background and changes in lighting will result in large errors. To overcome this, Yi et al. propose a Dual-mode SGM model.

\subsection{Dual-mode Single Gaussian Model}

In the Dual-mode SGM model, an absolute background model and a candidate background model are updated in concert. The absolute model represents the current model used for differentiating foreground from background. The candidate model represents a possible new background after, say, a shift in lighting or camera position. These models are updated as follows. Per pixel, if the squared absolute difference of the absolute model's mean and the pixel intensity is less than some weighted value of the variance, then the absolute model is updated as per the aforementioned SGM rules. This can be formulated as:

\begin{equation}
\left( M_{i}^{t} -  \mu_{A,i}^{(t)}\right)^2 < \theta_s \sigma_{A,i}^{(t)}
\end{equation}

If the absolute model is not updated, then the candidate model will be updated if it falls within the similarly defined threshold as follows:

\begin{equation}
\left( M_{i}^{t} -  \mu_{C,i}^{(t)}\right)^2 < \theta_s \sigma_{C,i}^{(t)}
\end{equation}

If neither model is updated, the candidate model is reset with the current mean set to the current pixel intensity, the variance set to some initial value (255 for the purposes of our implementation), and the age to 1.

If the age of the candidate model exceeds the age of the absolute model as in:

\begin{equation}
\alpha_{A,i}^{(t)} < \alpha_{C,i}^{(t)}
\end{equation}

the absolute model is set to the candidate model and the candidate model is reset.

\subsection{Motion Compensation}
The motion compensation described in Yi et al. is as follows. Kanade-Lucas-Tomasi's (KLT) feature tracking algorithm\footnote{KLT is a method of tracking the movement of pixels in an image sequence. In it, features (pixel sequences) are identified which are unique in an image. Methods based on Lucas-Kanade's optical flow are used to track these pixels through a series of images. Other methods similar to this include SURF~\cite{bay2006surf}.} is used to find feature points and the vector shifts of the pixels~\cite{lucas1981iterative, tomasi1991detection}. RANSAC~\cite{fischler1981random} is then used to create a homography matrix from these vectors. Because in their method, Yi et al. use grid blocks rather than single pixels, they compute the mixture between grid blocks based on this homography matrix. However, because we compute SGMs on a per-pixel basis, we can simply use the homography matrix to map the SGM values to the appropriate pixels based on the overall motion of the image.

\section{Methodology}

\subsection{Serial Implementation}

\subsubsection{Preprocessing}
For Gaussian and median blurring, we use two methods of implementation. First, we use the current OpenCV implementation of both. Second, we implement our own unoptimized methods for both. In the serial Gaussian blur, for each pixel, we multiply a 2D Gaussian kernel (created by the Gaussian equation) by the surrounding pixels. The sum of this is then placed in the current pixel's position in the updated image. For median blurring, we take the surrounding 8 pixels (though this can be changed, we generally stayed with 8), as well as the current pixel, and use a simple sorting algorithm to order their intensities. The median value is then used as the new value of the pixel. A snippet of the core serial code of these methods can be found in Appendices~\ref{App:preprocmediansnippet} and \ref{App:preprocgausssnippet}.

\subsubsection{Dual-Mode SGM}
As per the aforementioned equations described by Yi et al., we implement the Dual-mode SGM model as a core loop, keeping the age, variance, and mean of the two models as matrices. The core code iterated for each pixel can be seen in Appendix~\ref{App:dsgm}.

\subsubsection{Motion Compensation}
We implement motion compensation using OpenCV's implementation of KLT and RANSAC. We first use the findGoodFeatures function to generate a list of possible points. We then use KLT (by the method calcOpticalFlowPyrLK) to find the movement vectors associated with these points. Finally we feed these vectors into OpenCV's RANSAC (through the findHomography method). Lastly, we transform the current frame to match the previous frame's keypoints via this homography matrix to attempt to reduce error from overall camera motion. Snippets of this code can be found in Appendix~\ref{App:motion}. It is important to note that we did not achieve significant improvements from this method, which we will discuss in the results section. Because of the poor results and due to the efficiency of the existing OpenCV implementation and the complexity of the algorithms in comparison with the amount of time available for the project, we determined it to be out of the scope of the project to parallelize motion compensation as well as all other aspects.

\subsection{Thread Building Blocks Implementation}
Generally, for both the preprocessing parallelization and the Dual-mode SGM parallelization, we separate the inner process functions from the loops iterating over the pixels and formulate them as tasks with the overloaded TBB functions. We specify the number of tasks desired using the OpenCV TBB extension (cv::parallel\_for method). The image matrix is decomposed by rows based on the number of tasks allotted. TBB then parallelizes the tasks among multiple threads and processors.

\subsection{CUDA Implementation}

With CUDA, we used a general scheme of decomposition for the current image frame. We choose a number of threads per block to form square blocks ($N\times N$) and then we create a grid of blocks to match the size of the image. We provide CUDA with a simple kernel such that each thread will process a given pixel. We generally place the entire image into the global memory of the GPU, since past experimentation~\cite{so} seems to indicate that implementing a shared mechanism for the size of images we were using would have more overhead than a global memory implementation. The pixel that a thread worked on is determined according to standard CUDA conventions as:

\begin{Verbatim}[commandchars=\\\{\}]
		  \PYG{k}{const} \PYG{k+kt}{size\PYGZus{}t} \PYG{n}{r} \PYG{o}{=} \PYG{n}{blockIdx}\PYG{p}{.}\PYG{n}{x} \PYG{o}{*} \PYG{n}{blockDim}\PYG{p}{.}\PYG{n}{x} \PYG{o}{+} \PYG{n}{threadIdx}\PYG{p}{.}\PYG{n}{x}\PYG{p}{;}
		  \PYG{k}{const} \PYG{k+kt}{size\PYGZus{}t} \PYG{n}{c} \PYG{o}{=} \PYG{n}{blockIdx}\PYG{p}{.}\PYG{n}{y} \PYG{o}{*} \PYG{n}{blockDim}\PYG{p}{.}\PYG{n}{y} \PYG{o}{+} \PYG{n}{threadIdx}\PYG{p}{.}\PYG{n}{y}\PYG{p}{;}
		  \PYG{k}{const} \PYG{k+kt}{size\PYGZus{}t} \PYG{n}{index} \PYG{o}{=} \PYG{n}{r} \PYG{o}{*} \PYG{n}{numCols} \PYG{o}{+} \PYG{n}{c}\PYG{p}{;}
\end{Verbatim}

Here, r is the row of the pixel, c is the column of the pixel, and index becomes the position of the current pixel in an array based on the current Thread ID, Block Dimension, and Block ID (provided by CUDA libraries).

\subsubsection{Preprocessing}
To parallelize the Gaussian blur, we first attempted to simply write a CUDA kernel which would take as input the image and Gaussian kernel. The kernel would determine which pixel the current thread was assigned to through previously described means. Then the surrounding pixels would be convolved with this kernel. This however, is a naive implementation. In this scenario $O(nmk^2)$ operations are needed, where $n\times m$ is the size of the image, and $k\times k$ is the size of the kernel. To significantly reduce this cost, we can use a separable convolution. In this, a 1-Dimensional Gaussian kernel is used to convolve first the rows and then the columns~\cite{podlozhnyuk2007image}. This results in an equivalent output while requiring only $O(nm2k)$ operations. For the 2-Dimensional implementation of the Gaussian filter, we generally use the same code as the serial implementation in the CUDA kernel (obviously without looping over the pixels). For the separable convolution, we use similar code, but rather than running through a loop over a 2D Gaussian kernel for convolution we first run a CUDA kernel over the rows with a 1D Gaussian kernel, then we run it over the columns using the same code, with a flag raised. The core inner code remains fairly similar to the serial functions and can be found in the cuda\_kernels.cu file

For median blurring, we simply take the surrounding pixels, order them through a simple swap sorting mechanism and then choose the median value just as in the serial implementation. However, we divide this per kernel such that each thread calculates the values for only one pixel. Generally, the core of this code remains the same as the serial implementation.

\subsection{Dual-mode SGM}
To parallelize the Dual-mode SGM, we take the same inner code as the serial version and similarly remove the two outer loops which iterate over the pixels in favor of the CUDA kernel thread model described above.

\section{Results and Analysis}
For both qualitative and quantitative analysis, we use the University of Sherbrooke CDNet database~\cite{wang2014cdnet}. The database provides image sequences and videos for use in motion detection in different scenarios (including with motion of the camera).

\subsection{Qualitative Algorithm Analysis}
We first analyze the qualitative results of the algorithm. We note that all parallelized versions were verified against the serial version to ensure that no errors were introduced into the algorithm by parallelization. As such, this qualitative comparison is not necessary.

\subsubsection{Effects of Preprocessing}
\begin{figure}[h]
\centering
\begin{minipage}{.5\textwidth}
  \centering
  \includegraphics[width=.85\linewidth]{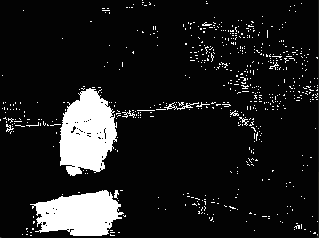}
  \captionof{figure}{Motion segmentation mask, no Gaussian or \\median filtering.}
  \label{fig:mask_nopre}
\end{minipage}%
\begin{minipage}{.5\textwidth}
  \centering
  \includegraphics[width=.85\linewidth]{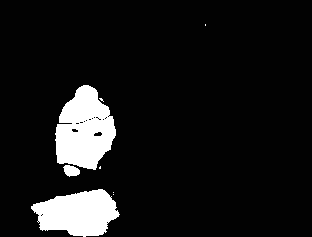}
  \captionof{figure}{Motion segmentation mask, Gaussian and median filtering (OpenCV implementation).}
  \label{fig:mask_pre}
\end{minipage}
\end{figure}

\begin{figure}[h]
\centering
\begin{minipage}{.5\textwidth}
  \centering
  \includegraphics[width=.85\linewidth]{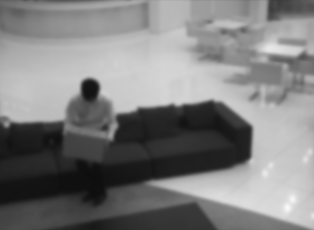}
  \captionof{figure}{Grey scale image, Gaussian and median \\filtering (OpenCV implementation).}
  \label{fig:grey_pre}
\end{minipage}%
\begin{minipage}{.5\textwidth}
  \centering
  \includegraphics[width=.85\linewidth]{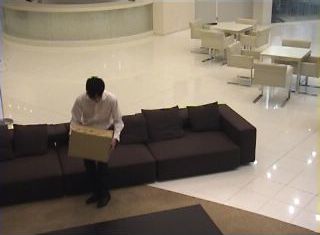}
  \captionof{figure}{This is the original image from the U. Sherbrooke CDNet database. The camera here remains static.}
  \label{fig:full_nopre}
\end{minipage}
\end{figure}

We determined that preprocessing has is an important aspect in filtering out noise during motion detection. This is clearly visible in Figures~\ref{fig:mask_nopre},\ref{fig:mask_pre},\ref{fig:grey_pre}, and \ref{fig:full_nopre}. Without preprocessing, the motion detection segmentation mask becomes grainy as more small anomalies in lighting are picked up on a per pixel basis. After Gaussian and median filtering, the anomalies disappear as the image smooths out.

\subsubsection{Comparison to OpenCV MOG}
\begin{figure}[h]
\centering
\begin{minipage}{.5\textwidth}
  \centering
  \includegraphics[width=.85\linewidth]{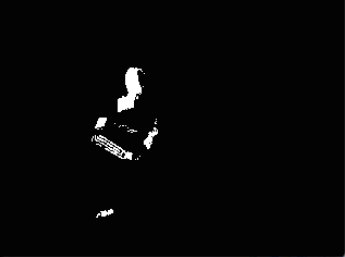}
  \captionof{figure}{OpenCV MoG algorithm implementation, \\we used all default parameters.}
  \label{fig:mog}
\end{minipage}%
\begin{minipage}{.5\textwidth}
  \centering
  \includegraphics[width=.85\linewidth]{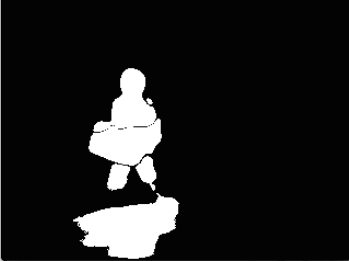}
  \captionof{figure}{The Dual-mode SGM algorithm, with preprocessing.}
  \label{fig:dsgm}
\end{minipage}
\end{figure}

Next, we wanted to compare our implementation of the Yi et al. algorithm against the Mixture of Gaussian model in OpenCV. MoG can potentially yield better results as it has a number of background models which can be used to develop a more robust representation of the background. In contrast to our expected results, we found that the MoG implementation of OpenCV did not produce readily identifiable objects in the image as seen in Figure~\ref{fig:mog}. Most of the man is considered to be part of the background, while the man is more identifiable in Figure~\ref{fig:dsgm} for our implementation. However, it is worth noting that any place where the color of the man's clothes blends in with his background, the algorithm cannot identify him as part of the foreground (as in his pants blending into the couch).

\begin{figure}[h]
\centering
\begin{minipage}{.5\textwidth}
  \centering
  \includegraphics[width=.85\linewidth]{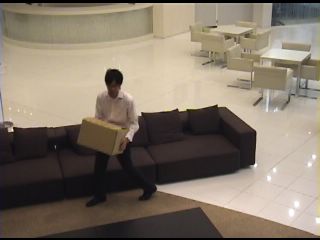}
  \captionof{figure}{The original image.}
  \label{fig:original_frame}
\end{minipage}%
\end{figure}

\subsubsection{Motion Compensation Results}

\begin{figure}[h]
\centering
\begin{minipage}{.5\textwidth}
  \centering
  \includegraphics[width=.85\linewidth]{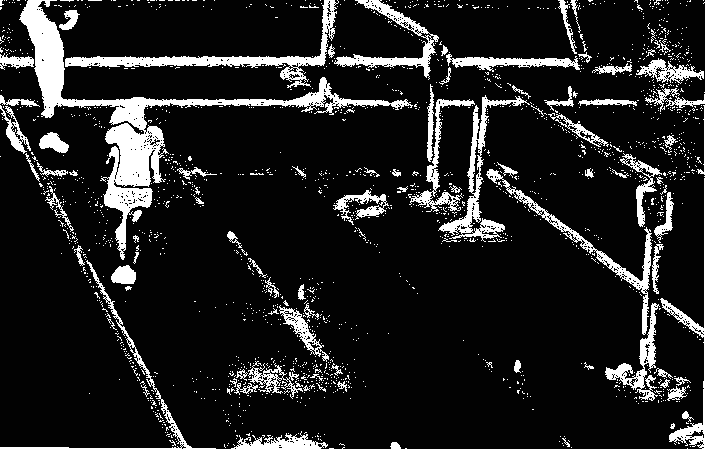}
  \captionof{figure}{The motion mask with motion \\compensation.}
  \label{fig:motioncomp}
\end{minipage}%
\begin{minipage}{.5\textwidth}
  \centering
  \includegraphics[width=.85\linewidth]{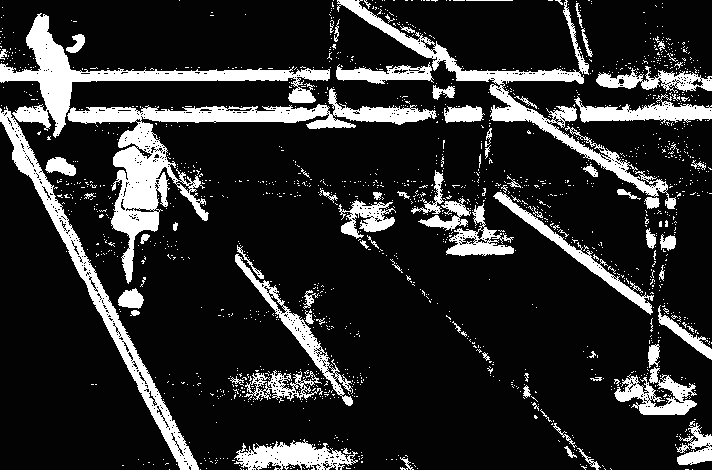}
  \captionof{figure}{The motion mask without motion \\compensation}
  \label{fig:nomotioncomp}
\end{minipage}
\end{figure}

\begin{figure}[h]
\centering
\begin{minipage}{.5\textwidth}
  \centering
  \includegraphics[width=.8\linewidth]{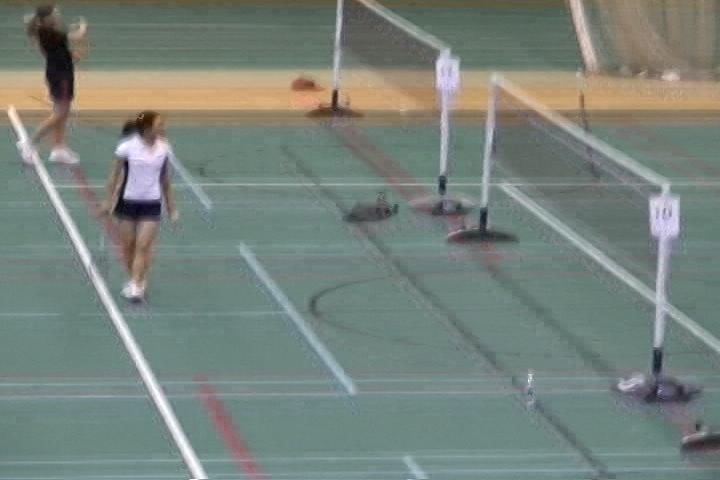}
  \captionof{figure}{The original image.}
  \label{fig:motioncomporig}
\end{minipage}%
\begin{minipage}{.5\textwidth}
  \centering
  \includegraphics[width=.905\linewidth]{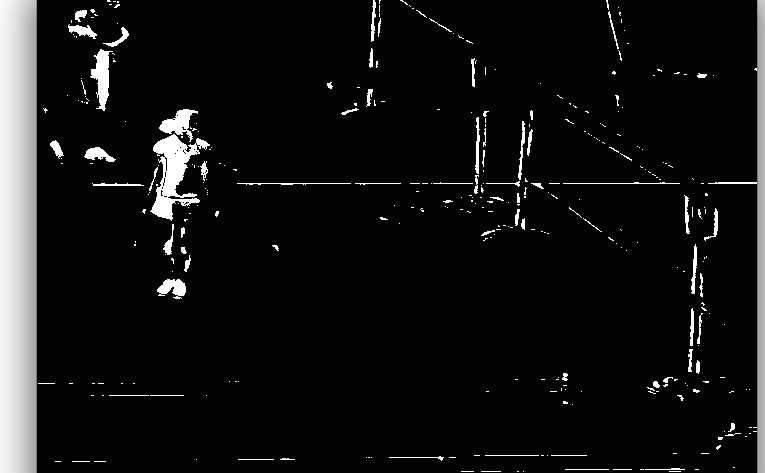}
  \captionof{figure}{The OpenCV MoG results with the algorithm.}
  \label{fig:motioncompmog}
\end{minipage}%
\end{figure}

While we implemented motion compensation using OpenCV's LKT, RANSAC, and transform functions, our results were not ideal and as such we did not focus on it for parallelization or other aspects. In a dataset with significant jitter as seen in Figures~\ref{fig:motioncomp}, \ref{fig:nomotioncomp}, and \ref{fig:motioncomporig}, we saw minimal benefit to having motion compensation versus not. As can be seen in Figure~\ref{fig:motioncompmog}, the MoG algorithm has a much cleaner image in this case. However, this is to be expected as it is both better suited to adaptable backgrounds and is much less sensitive than our implementation.

\subsection{Thread Building Blocks}
Originally, OpenMP was pursued for easy parallelization. However, on Mac OS X, compilation difficulties when building with OpenCV and OpenMP simultaneously were insurmountable. Since Thread Building Blocks (TBB) is readily supported on Mac OS X systems in combination with OpenCV and the compatible version of gcc, we use this.

\subsubsection{Test Specifications}
Combinations of using the OpenCV preprocessing filter versus an implemented version as well as including motion compensation in the computation yields four total variations: OpenCV preprocessing with Motion Compensation, OpenCV preprocessing without Motion Compensation, our preprocessing with Motion Compensation, and our preprocessing without Motion Compensation. Furthermore, serial and parallelized implementations of each variation were implemented.

An example set, showing a man enter the frame, sit on a sofa, and finally leave, was used. We use the first 500 frames of the set which has a resolution of 320x240. An original raw frame can be seen in Figure~\ref{fig:original_frame}. The test machine consisted of an Intel Core i5 at 2.4Ghz with two cores and four hardware threads.

\subsubsection{Results}
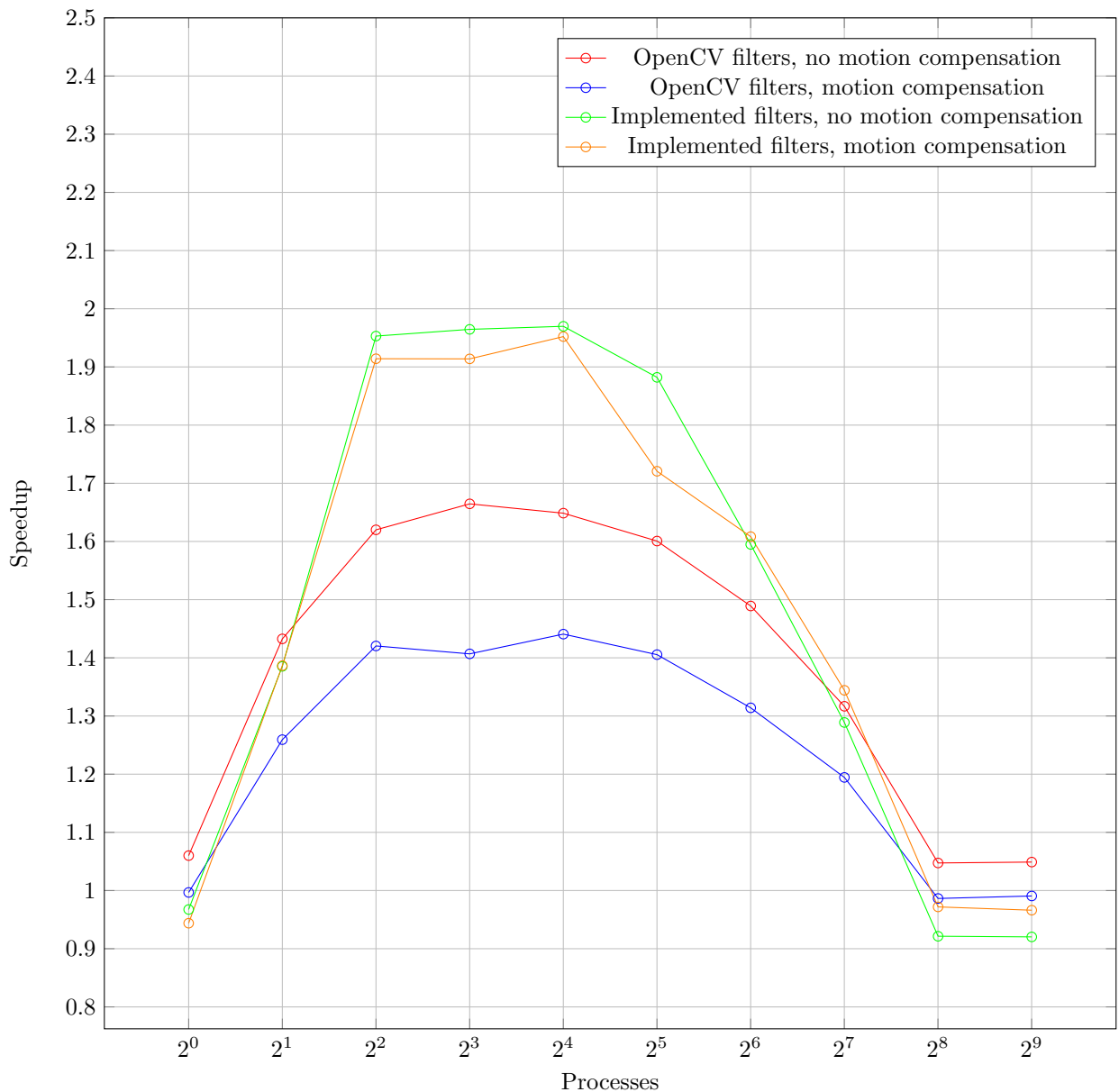
\begin{figure}[ht]
    \centering
\begin{tikzpicture}
\begin{axis}[height=1.0\linewidth,width=1.0\linewidth,xlabel=Processes ,ylabel=Speedup, xmode = log, log basis x=2, ymax=2.5, grid=major]
	\addplot[mark=o,color=red] table[x=num_threads,y=speedup] {speedup/speedup_ocvb_no_mot.dat};
	\addlegendentry{OpenCV filters, no motion compensation}
	\addplot[mark=o,color=blue] table[x=num_threads,y=speedup] {speedup/speedup_ocvb_mot.dat};
	\addlegendentry{OpenCV filters, motion compensation}
	\addplot[mark=o,color=green] table[x=num_threads,y=speedup] {speedup/speedup_no_ocvb_no_mot.dat};
	\addlegendentry{Implemented filters, no motion compensation}
	\addplot[mark=o,color=orange] table[x=num_threads,y=speedup] {speedup/speedup_no_ocvb_mot.dat};
	\addlegendentry{Implemented filters, motion compensation}
\end{axis}
\end{tikzpicture}
\caption{Speedup comparison of the four main variations of DSGM. An example set of 500 frames as 320x240 was tested on a machine with an Intel Core i5 at 2.4Ghz with two cores and four threads. The use of OpenCV filters over the implemented filters typically speeds computation due to optimizations in the OpenCV library. Since the running time of the OpenCV filters is constant across all implementations, it does not contribute to the speedup. Using the implement filters, however, greatly contributes to the speedup of a variation since there is a significant running time difference between the serial and parallel implementations of the filters. Similar to the OpenCV filters, motion detection is constant time processing and thus diminishes the speed of a variation.}
    \label{fig:speedup_tbb}
\end{figure}

\begin{figure}[ht]
    \centering
\begin{tikzpicture} 
	\begin{axis}[ybar stacked, bar width=10pt, width=1.0\linewidth,xlabel=Tasks,ylabel=Time (seconds), symbolic x coords={Serial,1,2,4,8,16,32,64,128,256,512}, xtick=data, x tick label style={rotate=0,anchor=north}, enlarge x limits=0.12,
enlarge y limits=0.00, ymax=130, ymin=0]
	\addplot table[x=num_threads,y=t_blur] {hist/no_ocvb_no_mot.dat};
	\addlegendentry{Preprocessing}
	\addplot table[x=num_threads,y=t_dsgm] {hist/no_ocvb_no_mot.dat};
	\addlegendentry{DSGM update}
	\addplot table[x=num_threads,y=t_serl] {hist/no_ocvb_no_mot.dat};
	\addlegendentry{Other serial}
	\end{axis}
\end{tikzpicture}
\caption{Using an implemented Gaussian and median filter with motion compensation disabled. An example set of 500 frames as 320x240 was tested on a machine with an Intel Core i5 at 2.4Ghz with two cores and four hardware threads. The implemented filter accounts for the majority of the processing time even with TBB. Despite the slow running time, the effectiveness of parallelization with TBB is evident.}
    \label{fig:no_ocvb_no_mot}
\end{figure}
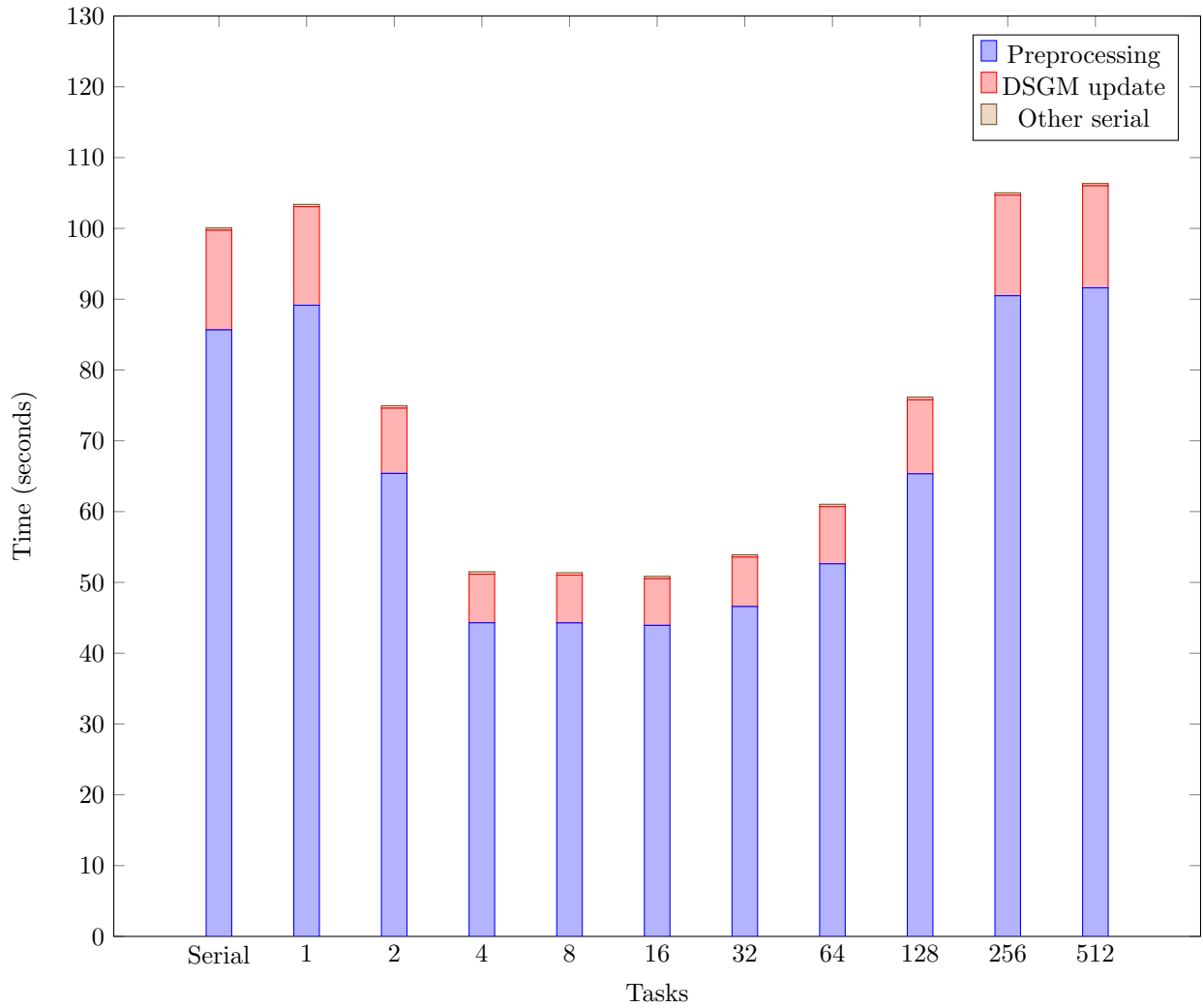

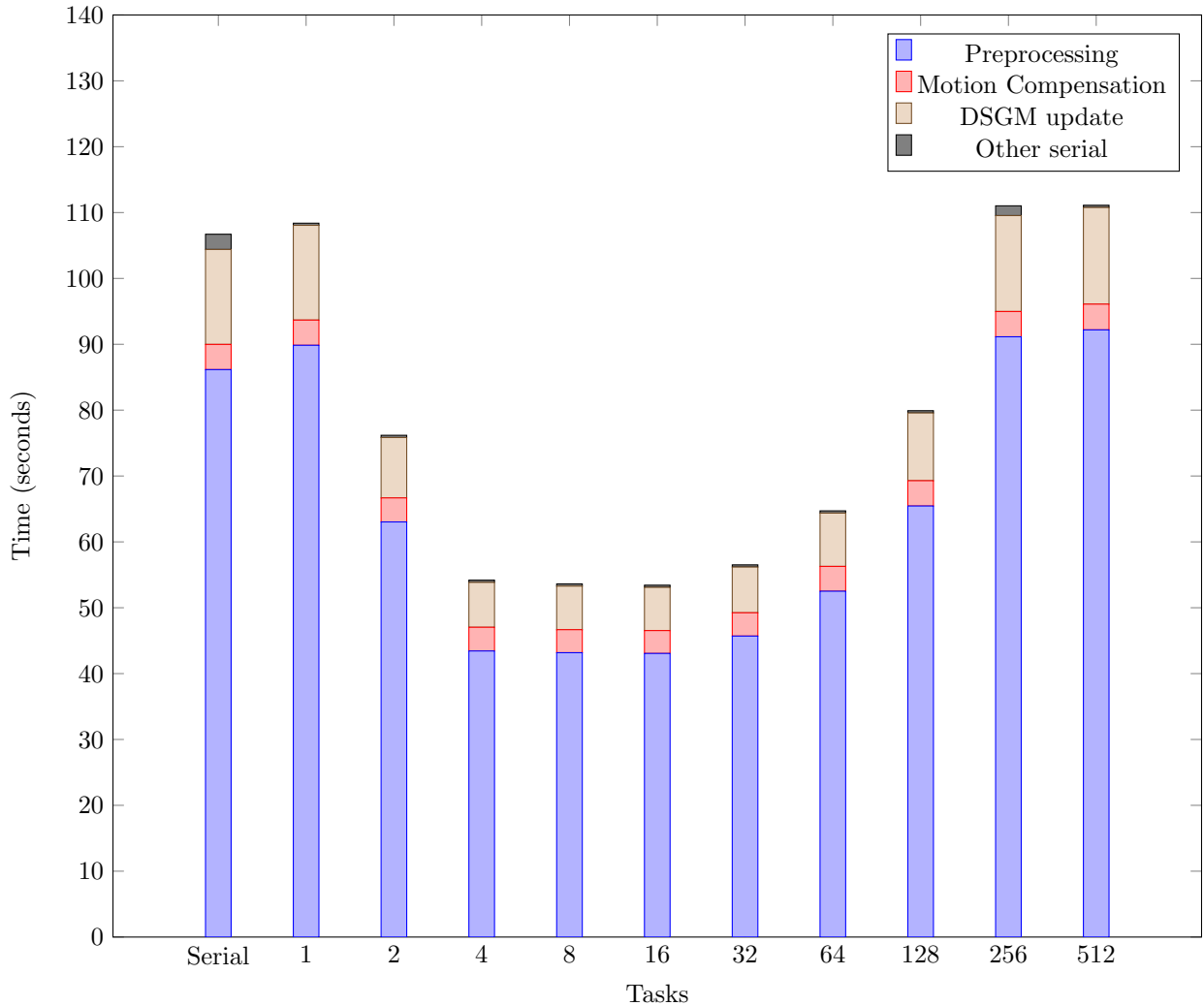
\begin{figure}[ht]
    \centering
\begin{tikzpicture} 
	\begin{axis}[ybar stacked, bar width=10pt, width=1.0\linewidth,xlabel=Tasks,ylabel=Time (seconds), symbolic x coords={Serial,1,2,4,8,16,32,64,128,256,512}, xtick=data, x tick label style={rotate=0,anchor=north}, enlarge x limits=0.12,
enlarge y limits=0.00, ymax=140, ymin=0]
	\addplot table[x=num_threads,y=t_blur] {hist/no_ocvb_mot.dat};
	\addlegendentry{Preprocessing}
	\addplot table[x=num_threads,y=t_mtnc] {hist/no_ocvb_mot.dat};
	\addlegendentry{Motion Compensation}
	\addplot table[x=num_threads,y=t_dsgm] {hist/no_ocvb_mot.dat};
	\addlegendentry{DSGM update}
	\addplot table[x=num_threads,y=t_serl] {hist/no_ocvb_mot.dat};
	\addlegendentry{Other serial}
	\end{axis}
\end{tikzpicture}
\caption{Using an implemented Gaussian and median filter with motion compensation enabled for 500 frames as 320x240. The test machine used a Intel Core i5 at 2.4Ghz with two cores and four hardware threads. The implemented filter accounts for the majority of the processing time even with TBB. Despite the slow running time, the effectiveness of parallelization with TBB is evident. With motion compensation enabled, a steady serial processing time is added to each test.}
    \label{fig:no_ocvb_mot}
\end{figure}

Figure~\ref{fig:speedup_tbb} displays the speedup of each the four parallel variations as compared to their respective serial versions with a varying number of TBB tasks. Figures~\ref{fig:ocvb_no_mot},~\ref{fig:ocvb_mot},~\ref{fig:no_ocvb_no_mot} and~\ref{fig:no_ocvb_mot} reveals the time breakdown spent in each of the major parts of the complete algorithm for all four variations. It is important to note that while we can specify tasks to be parallelized TBB by default creates threads optimized for the number of processor cores and hardware threads. TBB then schedules these tasks onto the cores and hardware threads~\cite{kim2011multicore}. As such, we speak of tasks, not threads.

Across all the variations, there are a number of general trends. The optimal number of TBB tasks for running the algorithm plateaus between four and sixteen. Since the test machine has four hardware threads, this is a consistent result. TBB can schedule all the tasks successfully among the hardware threads. Comparing the runtime of the serial version to a single TBB task reveals the parallelization overhead associated with TBB since the serial version is typically slight faster. Furthermore, specifying TBB to use a number of tasks greater than 256 results in times comparable to only using one task. An unreasonably high number of tasks results in the framework not being able to schedule the task among multiple processors or hardware threads, and hence the result is the same as having only one thread. At this point the tasks become to small so the first core is complete with a task before the library can schedule the next one on a different processor so it schedules it on the same thread on the same core~\cite{kim2011multicore}. The addition of motion compensation into an algorithm delivers a near constant time addition to the serial portion of the runtime with little variation.

With regards to using the OpenCV Gaussian and median filters for preprocessing as compared to the serial and parallelized filters implemented in this project, the OpenCV implementations are consistently much faster. Despite any TBB parallelization used to speed up filtering, the OpenCV image processing library is optimized based on the architecture of the processor used~\footnote{The optimizations within OpenCV were revealed by examining the source code found on GitHub. Many preprocessing flags for specific processor architectures are used, indicating hardware specific optimizations. In particular: \url{https://github.com/Itseez/opencv/blob/f50f249f804b35ad6f39459ec6df9a66fb0825f8/modules/imgproc/src/smooth.cpp}}. Ultimately, any attempt to re-implement and parallelize parts of OpenCV would most likely be significantly slower.

\begin{figure}[ht]
    \centering
\begin{tikzpicture} 
	\begin{axis}[ybar stacked, bar width=10pt, width=1.0\linewidth,xlabel=Tasks,ylabel=Time (seconds), symbolic x coords={Serial,1,2,4,8,16,32,64,128,256,512}, xtick=data, x tick label style={rotate=0,anchor=north}, enlarge x limits=0.12,
enlarge y limits=0.00, ymax=28, ymin=0]
	\addplot table[x=num_threads,y=t_blur] {hist/ocvb_no_mot.dat};
	\addlegendentry{Preprocessing}
	\addplot table[x=num_threads,y=t_dsgm] {hist/ocvb_no_mot.dat};
	\addlegendentry{DSGM update}
	\addplot table[x=num_threads,y=t_serl] {hist/ocvb_no_mot.dat};
	\addlegendentry{Other serial}
	\end{axis}
\end{tikzpicture}
\caption{Using OpenCV Gaussian and median filtering with motion compensation disabled. An example set of 500 frames as 320x240 was tested on a machine with an Intel Core i5 at 2.4Ghz with two cores and four hardware threads. Across all tests, the OpenCV filtering remains constant time. The TBB parallelization shows faster running times with 4-16 tasks.}
    \label{fig:ocvb_no_mot}
\end{figure}
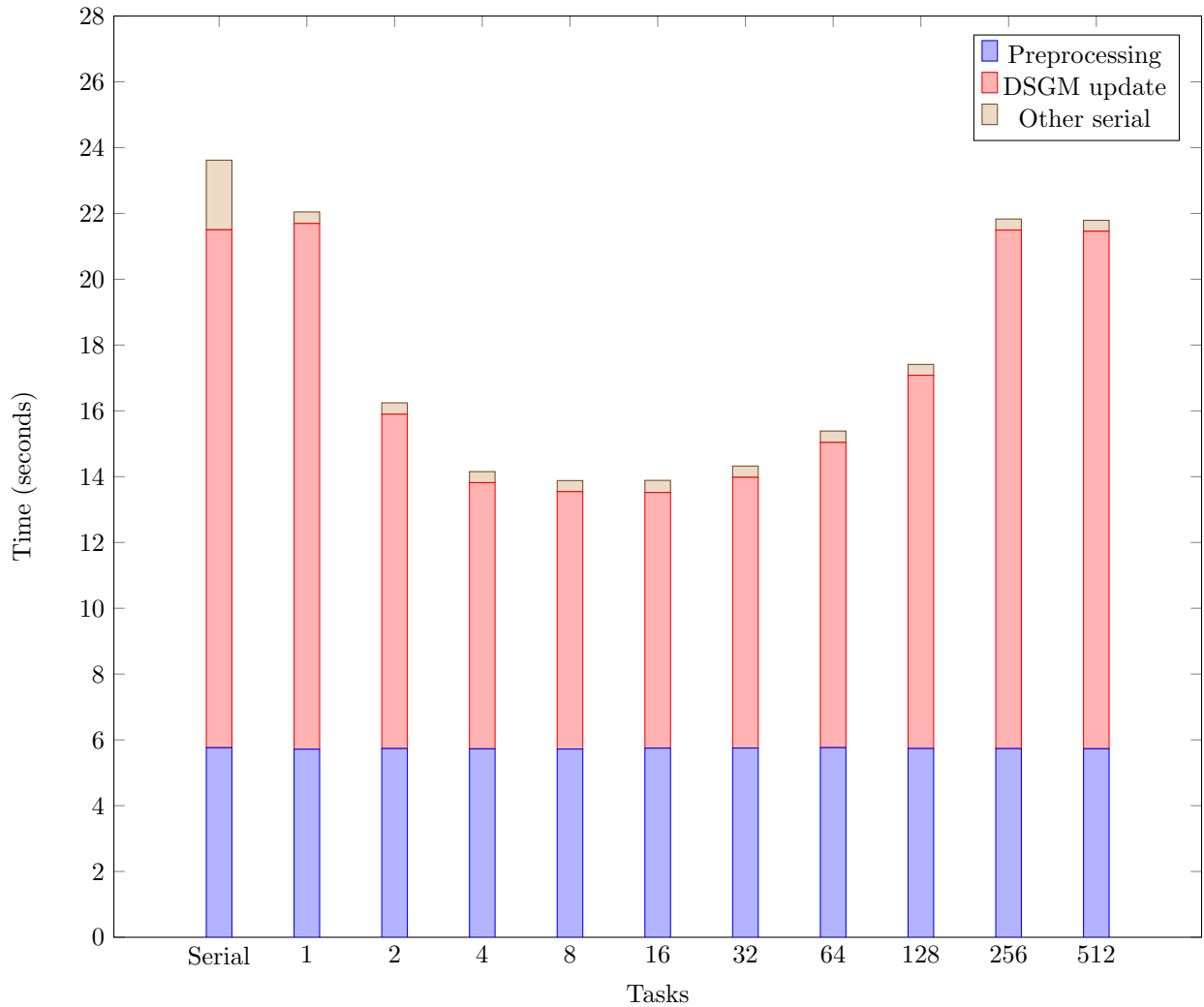

\begin{figure}[ht]
    \centering
\begin{tikzpicture} 
	\begin{axis}[ybar stacked, bar width=10pt, width=1.0\linewidth,xlabel=Tasks,ylabel=Time (seconds), symbolic x coords={Serial,1,2,4,8,16,32,64,128,256,512}, xtick=data, x tick label style={rotate=0,anchor=north}, enlarge x limits=0.12,
enlarge y limits=0.00, ymax=32, ymin=0]
	\addplot table[x=num_threads,y=t_blur] {hist/ocvb_mot.dat};
	\addlegendentry{Preprocessing}
	\addplot table[x=num_threads,y=t_mtnc] {hist/ocvb_mot.dat};
	\addlegendentry{Motion Compensation}
	\addplot table[x=num_threads,y=t_dsgm] {hist/ocvb_mot.dat};
	\addlegendentry{DSGM update}
	\addplot table[x=num_threads,y=t_serl] {hist/ocvb_mot.dat};
	\addlegendentry{Other serial}
	\end{axis}
\end{tikzpicture}
\caption{Using OpenCV Gaussian and median filtering with motion compensation enabled. An example set of 500 frames as 320x240 was tested on a machine with an Intel Core i5 at 2.4Ghz with two cores and four hardware threads. Across all tests, the OpenCV filtering remains constant time. With motion compensation enabled, a steady serial processing time is added to each test.}
    \label{fig:ocvb_mot}
\end{figure}
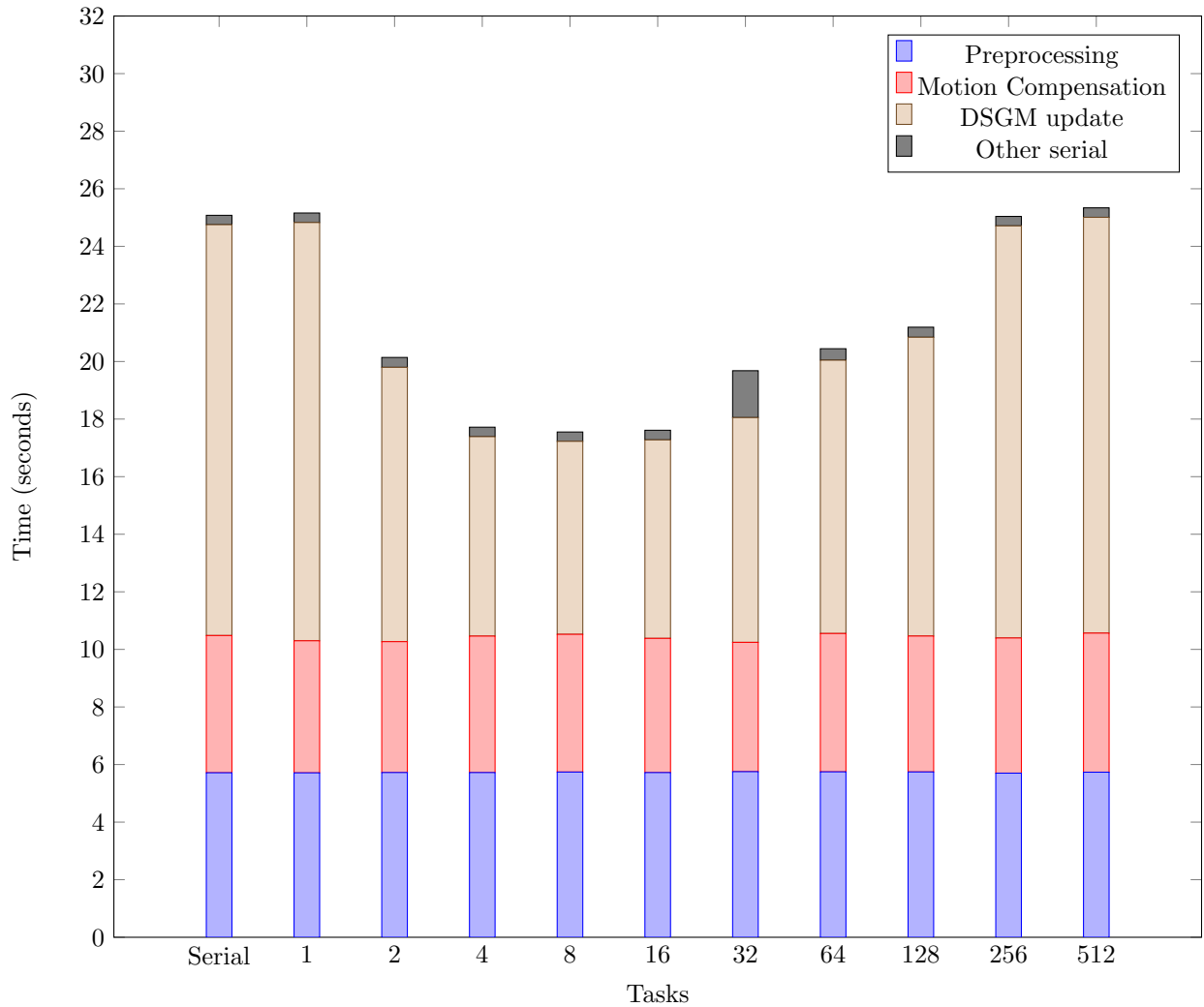

\subsection{CUDA}

\subsubsection{Machine Specifications}
The main graphics card we use for testing is an NVIDIA GeForce GT 650M graphics card in conjunction with a 2.6 GHz Intel Core i7 processor. Though full specifications for the card from CUDA's deviceQuery program can be seen in Appendix~\ref{App:device_specs}, the main points to note are the following. It is a Kepler architecture card with two multiprocessors with 192 CUDA cores per multiprocessor. It has a warp size of 32 and a quad-warp scheduler (it schedules 4 warps at once for a total of 128 threads). Additionally it can handle a maximum of 1024 threads per block.

\subsubsection{Preprocessing}
We first compared our separable Gaussian filter implementation to the 2-Dimensional implementation. We ran this test on the NVIDIA GeForce GT 650M graphics card. We excluded motion compensation, but retained median blurring and the CUDA-parallelized Dual-mode SGM background detection for all versions that we ran. We found as we previously did with TBB that OpenCV Gaussian and Median filters are significantly faster overall due the heavy optimizations done within the code. Additionally, though the CUDA parallelized version of separable Gaussian filtering comes closer to comparing to the OpenCV method, but still cannot compete with its optimizations. As such, for further tests we retain the OpenCV pre-processing methods.

\begin{figure}[h]
\centering\begin{tabular}{ |c|c|c|c| }
	\hline
  \pbox{20cm}{Our Serial Gaussian Filter\\and Median Filter} & \pbox{20cm}{.\\OpenCV Serial \\Gaussian Filter \\and Median Filter\\} & \pbox{20cm}{CUDA 2-D Implementation\\ and CUDA Median Filter} & \pbox{20cm}{CUDA Separable Filter \\and CUDA Median Filter}\\
  \hline
  50.74 seconds & 9.35 seconds & 18.26 seconds & 13.03 seconds \\
  \hline
\end{tabular}
  \caption{A comparison of CUDA 2-Dimensional and separable Gaussian filter implementation runtimes to our serial Gaussian Filter and OpenCV's serial Gaussian filter. Note: these numbers were taken from the average of 20 runs using an NVIDIA GeForce GT 650M graphics card and a 2.6 GHz Intel Core i7 processor. Includes CUDA speedup of all other parts except the filtering. Excludes motion compensation altogether. 121 threads per block for CUDA kernels.}
\end{figure}

\subsubsection{Dual-mode SGM and Thread Block Optimization}
\begin{figure}[h]
\hspace*{-2cm}
\centering
\begin{tikzpicture}
\begin{axis}[ybar stacked, bar width=5pt, enlargelimits=0.08, width=1\linewidth, height=8cm, xlabel=Threads per block,ylabel=Time (seconds), symbolic x coords={1,4,9,16,25,36,49,64,81,100,121,144,169,196,225,256,289,324,361,400,441,484,529,576,625,676,729,784,841,900,961,1024} ,xtick=data, x tick label style={rotate=45,anchor=east}, every axis/.append style={font=\small},]
\addplot table[x=P,y=t_ser] {hist/cuda.dat};
\addlegendentry{Serial}
\addplot table[x=P,y=t_proc] {hist/cuda.dat};
\addlegendentry{Parallel Processing}
\addplot table[x=P,y=t_comm] {hist/cuda.dat};
\addlegendentry{Parallel Communication}
\end{axis}
\end{tikzpicture}
\caption{Here we see the proportional parts of the code which run in Serial and Parallel as well as the breakdown of parallel communication time vs. parallel processing time. Communication and processing was determined as follows. A CPU clock was used to measure the total time in parallel (including memory operations). Subsequently, the GPU clock (through CUDA library event calls) was used to determine to amount of time spent processing on the GPU. The two were subtracted to gain insight into parallel vs. computation time. Test run on NVIDIA GeForce GT 650M graphics card.}
\label{fig:time_cuda}
\end{figure}
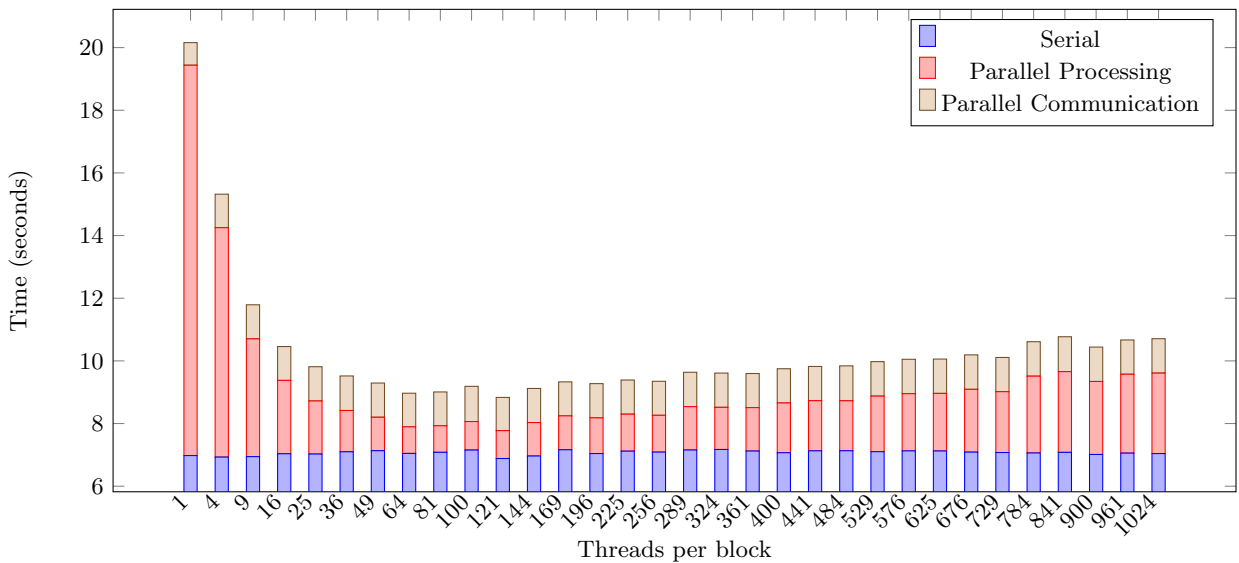

Overall, by parallelizing the Dual-Mode SGM we see significant speedup from the serial version of our code. To test the speedup for this part, we use an OpenCV Gaussian Filter and Median Filter and exclude motion compensation. Furthermore, we attempt to demonstrate thread block size optimization on the NVIDIA Kepler architecture with CUDA. As seen in Figures~\ref{fig:speedupCUDA} and~\ref{fig:time_cuda}, we attain a significant maximal speedup of about 2.5 using CUDA parallelization with 121 threads per block (a square block of size $N\times N$). As seen in Figure~\ref{fig:time_cuda}, where we consider communication time of parallel processing to be transfer of data between GPU memory and main memory, for all block size the communication time remains constant, as we only copy over the image once to GPU memory. Processing time, however, significantly drops to an optimal speedup at 121 threads. This correlates greatly with expectations. In the Kepler architecture and the NVIDIA GeForce GT 650M graphics card, the warp size is 32 with a quad-warp scheduler. That means that 128 threads are scheduled at once. At 121 threads, nearly all warps are filled and scheduled at once as soon as a block is brought into the multiprocessor. This is the reason we expect this to be an optimal thread block size. With more threads, however, all warps may not be scheduled or made use of at once and hence leaving some processors idle. This is why we see decreases in speedup both with more threads and less threads per block. It is important to note that for this test we displayed each frame using the OpenCV imshow function, as such there is a larger serial overhead.

\begin{figure}[h]
\centering
\begin{tikzpicture}
\tikzstyle{every node}=[font=\small]
\begin{axis}[width=0.9\linewidth,xlabel=Threads per block,ylabel=Speedup, symbolic x coords={1,4,9,16,25,36,49,64,81,100,121,144,169,196,225,256,289,324,361,400,441,484,529,576,625,676,729,784,841,900,961,1024} ,xtick=data, x tick label style={rotate=45,anchor=east}, grid=major]
	\addplot[mark=o,color=red] table[x=P,y=speedup] {speedup/speedupCUDA.dat};
	\addlegendentry{CUDA}
\end{axis}
\end{tikzpicture}
\caption{Here we see the speedup of the CUDA implementation when compared to the unparallelized version. Both use the OpenCV preprocessing methods. Tests run on NVIDIA GeForce GT 650M graphics card.}
\label{fig:speedupCUDA}
\end{figure}

\subsubsection{GPU Comparison}
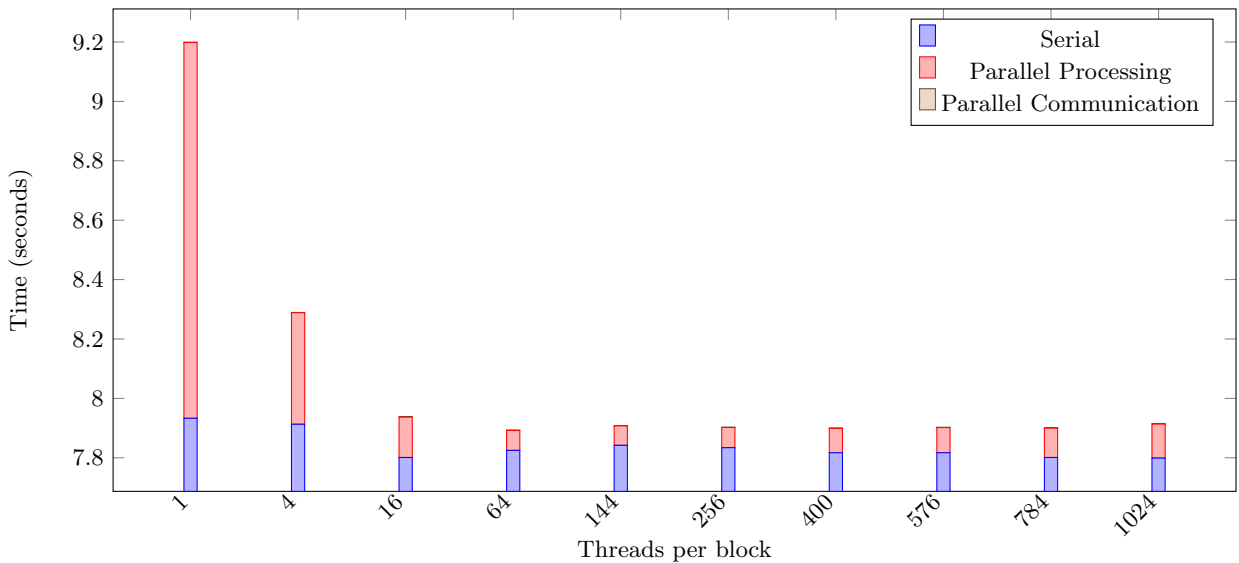
\begin{figure}[h]
\hspace*{-2cm}
\centering
\begin{tikzpicture}
\begin{axis}[ybar stacked, bar width=5pt, enlargelimits=0.08, width=1\linewidth, height=8cm, xlabel=Threads per block,ylabel=Time (seconds), symbolic x coords={1,4,16,64,144,256,400,576,784,1024} ,xtick=data, x tick label style={rotate=45,anchor=east}, every axis/.append style={font=\small},]
\addplot table[x=P,y=t_ser] {hist/cuda_gtx570.dat};
\addlegendentry{Serial}
\addplot table[x=P,y=t_proc] {hist/cuda_gtx570.dat};
\addlegendentry{Parallel Processing}
\addplot table[x=P,y=t_comm] {hist/cuda_gtx570.dat};
\addlegendentry{Parallel Communication}
\end{axis}
\end{tikzpicture}
\caption{Here we see the same test as the original CUDA tests ran previously with varying threads per block on the GTX 570 card. We use a smaller subset of tests and see little variation across the number of threads per block. We used a Windows machine in this case so the timing information only goes to 3 decimal places for CPU based timing due to the Windows timing library. Communication is negligible at a value of 0.001 seconds.}
\label{fig:time_cuda2}
\end{figure}

We also acquired another GPU for comparison under the Fermi architecture. The second card used was a GeForce GTX 570 with full specifications found in Appendix~\ref{App:device_specs2}. The main difference between the two cards (emphasized due to the differences in architecture) is the number of multiprocessors and CUDA cores in this one. There are 15 multiprocessors with 32 CUDA Cores each for a total of 480 CUDA Cores. The warp size is also 32, but there is only a dual warp scheduler. We ran the same tests as on the previous card with various amounts of threads per block (though a smaller subset) which can be found in Figures~\ref{fig:speedupCUDAgtx} and~\ref{fig:time_cuda2}. Here it is visible that we reach a speedup wall with this processor. While there is some improvement from 1 thread per block to 64 threads per block. The serial portion of the code (OpenCV preprocessing, which showed to be faster than our parallelized preprocessing, and displaying of images through the OpenCV library) proved to be much more overhead than the parallel processing. As such we see a speedup of about 2.7 at maximum. Additionally, because the GTX 570 has more blocks scheduled at once (among 15 multiprocessors, so 15 blocks at once) with two warp schedulers of 32 threads each. It makes sense that 64 threads per block would be the most efficient. As a block is scheduled all threads are immediately scheduled by the two warp schedulers and 15 blocks can be handled almost simultaneously. However, the overall time of processing on the GPU is nearly negligible and hence speedup does not change drastically for varying numbers of threads (except for 1 and 4 threads per block, in which case too many CUDA cores are wasted during scheduling of a warp). Because this is a real time algorithm and the output of the image takes more time than the processing on a GPU with a reasonable thread block size, running it on more powerful GPUs is unnecessary. As it is, the algorithm has become real time.

\begin{figure}[h]
\centering
\begin{tikzpicture}
\tikzstyle{every node}=[font=\small]
\begin{axis}[width=0.9\linewidth,xlabel=Threads per block,ylabel=Speedup, symbolic x coords={1,4,16,64,144,256,400,576,784,1024} ,xtick=data, x tick label style={rotate=45,anchor=east}, grid=major]
	\addplot[mark=o,color=red] table[x=P,y=speedup] {speedup/speedupCUDAgtx570.dat};
	\addlegendentry{CUDA}
\end{axis}
\end{tikzpicture}
\caption{Here we see the speedup of the CUDA implementation when compared to the unparallelized version. Both use the OpenCV preprocessing methods. Tests run on NVIDIA GeForce GTX 570 graphics card.}
\label{fig:speedupCUDAgtx}
\end{figure}

\section{Conclusion}
We successfully implemented the method in Yi et al. for detecting and segmenting motion in a series of images\footnote{Our implementation can be found at: \url{https://github.com/Breakend/MotionDetection}.}. While we had poor results in achieving motion compensation, the results with a static camera are promising and show more detail than OpenCV's MoG implementation. We successfully achieved speedup with both CUDA and TBB, however, for preprocessing, OpenCV's implementation was much more optimized and thus resulted in no improvement from the OpenCV implementation in overall running time. Additionally, on a somewhat powerful GPU like the GTX 570 with many multiprocessors (per the Fermi architecture), we see that the speedup of the algorithm through parallelization hits a wall. The time to read and display an image using OpenCV functions (and the slight time for preprocessing) overtakes all other functionality. As such, this is essentially a real time algorithm.

\subsection{Issues and Usability of TBB}
While we chose TBB for its ease of installation and lack of compilation issues with our systems (Macbook Pro laptops with Intel processors running OS X 10.9), it is helpful in its abstraction of many small optimizations relating to multiprocessor use. Once the learning curve of TBB was surpassed, it was relatively simple to formulate different functions as tasks for the library to divide among processors.

\subsection{Issues and Usability of CUDA}
CUDA is an extremely helpful tool in parallelization with the GPU, we achieved significant speedup using the tool. While more speedup may be possible by optimizing distribution of global memory into shared memory caches, we found previous work that indicates this would not be helpful with the size of images we are using (as previously mentioned). CUDA itself is a tool that is difficult to debug, so issues (including with invalid memory corruption in the case of non-contiguously allocated matrices) took a long time to trace. Overall, CUDA makes GPU programming more accessible, assuming you are able to successfully install and compile it. The abstractions of thread blocks make parallelization relatively simple once you have some core base code complete. However, the learning curve for CUDA is steep.

\subsection{Future Work}
In the future, we hope to optimized and modify Yi et al.'s method to work with more robust motion compensation systems to improve and parallelize those results. We also hope to implement some parallelizations of the basic algorithm on the Raspberry Pi's modest GPU so that low cost mobile systems (as in robotics) are possible.

\clearpage
\addcontentsline{toc}{section}{References}
\bibliographystyle{IEEEtran}
\bibliography{references}

\clearpage
\begin{appendices}

\section{Device Specifications: GeForce GT 650M} \label{App:device_specs}

		\begin{Verbatim}[commandchars=\\\{\},codes={\catcode`$=3\catcode`_=8}, fontsize=\small]
Device 0: "GeForce GT 650M"
  CUDA Driver Version / Runtime Version          6.5 / 6.5
  CUDA Capability Major/Minor version number:    3.0
  Total amount of global memory:                 \textbf{1024 MBytes} (1073414144 bytes)
  ( 2) Multiprocessors, (192) CUDA Cores/MP:     \textbf{384 CUDA} Cores
  GPU Clock rate:                                \textbf{405 MHz} (0.41 GHz)
  Memory Clock rate:                             2000 Mhz
  Memory Bus Width:                              128-bit
  L2 Cache Size:                                 262144 bytes
  Maximum Texture Dimension Size (x,y,z)         1D=(65536), 2D=(65536, 65536), 3D=(4096, 4096, 4096)
  Maximum Layered 1D Texture Size, (num) layers  1D=(16384), 2048 layers
  Maximum Layered 2D Texture Size, (num) layers  2D=(16384, 16384), 2048 layers
  Total amount of constant memory:               65536 bytes
  Total amount of shared memory per block:       49152 bytes
  Total number of registers available per block: 65536
  Warp size:                                     \textbf{32}
  Maximum number of threads per multiprocessor:  2048
  Maximum number of threads per block:           \textbf{1024}
  Max dimension size of a thread block (x,y,z): (1024, 1024, 64)
  Max dimension size of a grid size    (x,y,z): (2147483647, 65535, 65535)
  Maximum memory pitch:                          2147483647 bytes
  Texture alignment:                             512 bytes
  Concurrent copy and kernel execution:          Yes with 1 copy engine(s)
  Run time limit on kernels:                     Yes
  Integrated GPU sharing Host Memory:            No
  Support host page-locked memory mapping:       Yes
  Alignment requirement for Surfaces:            Yes
  Device has ECC support:                        Disabled
  Device supports Unified Addressing (UVA):      Yes
  Device PCI Bus ID / PCI location ID:           1 / 0
		\end{Verbatim}

\section{Device Specifications: GeForce GTX 570}
\label{App:device_specs2}
		\begin{Verbatim}[commandchars=\\\{\},codes={\catcode`$=3\catcode`_=8}, fontsize=\small]

Device 0: "GeForce GTX 570"
  CUDA Driver Version / Runtime Version          6.5 / 6.5
  CUDA Capability Major/Minor version number:    2.0
  Total amount of global memory:                 1280 MBytes (1342177280 bytes)
  (15) Multiprocessors, ( 32) CUDA Cores/MP:     480 CUDA Cores
  GPU Clock rate:                                1464 MHz (1.46 GHz)
  Memory Clock rate:                             1900 Mhz
  Memory Bus Width:                              320-bit
  L2 Cache Size:                                 655360 bytes
  Maximum Texture Dimension Size (x,y,z)         1D=(65536), 2D=(65536, 65535), 3D=(2048, 2048, 2048)
  Maximum Layered 1D Texture Size, (num) layers  1D=(16384), 2048 layers
  Maximum Layered 2D Texture Size, (num) layers  2D=(16384, 16384), 2048 layers
  Total amount of constant memory:               65536 bytes
  Total amount of shared memory per block:       49152 bytes
  Total number of registers available per block: 32768
  Warp size:                                     32
  Maximum number of threads per multiprocessor:  1536
  Maximum number of threads per block:           1024
  Max dimension size of a thread block (x,y,z): (1024, 1024, 64)
  Max dimension size of a grid size    (x,y,z): (65535, 65535, 65535)
  Maximum memory pitch:                          2147483647 bytes
  Texture alignment:                             512 bytes
  Concurrent copy and kernel execution:          Yes with 1 copy engine(s)
  Run time limit on kernels:                     Yes
  Integrated GPU sharing Host Memory:            No
  Support host page-locked memory mapping:       Yes
  Alignment requirement for Surfaces:            Yes
  Device has ECC support:                        Disabled
  CUDA Device Driver Mode (TCC or WDDM):         WDDM (Windows Display Driver Model)
  Device supports Unified Addressing (UVA):      Yes
  Device PCI Bus ID / PCI location ID:           1 / 0
\end{Verbatim}

\section{Preprocessing Gaussian Blur Code Snippet}
\label{App:preprocgausssnippet}
\begin{Verbatim}[commandchars=\\\{\}]
\PYG{k+kt}{float} \PYG{n}{blur} \PYG{o}{=} \PYG{l+m+mf}{0.f}\PYG{p}{;}
\PYG{c+c1}{//Average pixel color summing up adjacent pixels.}
\PYG{k}{for} \PYG{p}{(}\PYG{k+kt}{int} \PYG{n}{i} \PYG{o}{=} \PYG{o}{\PYGZhy{}}\PYG{n}{half}\PYG{p}{;} \PYG{n}{i} \PYG{o}{\PYGZlt{}=} \PYG{n}{half}\PYG{p}{;} \PYG{o}{++}\PYG{n}{i}\PYG{p}{)} \PYG{p}{\PYGZob{}}
  \PYG{k}{for} \PYG{p}{(}\PYG{k+kt}{int} \PYG{n}{j} \PYG{o}{=} \PYG{o}{\PYGZhy{}}\PYG{n}{half}\PYG{p}{;} \PYG{n}{j} \PYG{o}{\PYGZlt{}=} \PYG{n}{half}\PYG{p}{;} \PYG{o}{++}\PYG{n}{j}\PYG{p}{)} \PYG{p}{\PYGZob{}}
	\PYG{c+c1}{// Clamp filter to the image border}
	\PYG{k+kt}{int} \PYG{n}{h} \PYG{o}{=} \PYG{n}{min}\PYG{p}{(}\PYG{n}{max}\PYG{p}{(}\PYG{n}{r} \PYG{o}{+} \PYG{n}{i}\PYG{p}{,} \PYG{l+m+mi}{0}\PYG{p}{),} \PYG{n}{height}\PYG{p}{);}
	\PYG{k+kt}{int} \PYG{n}{w} \PYG{o}{=} \PYG{n}{min}\PYG{p}{(}\PYG{n}{max}\PYG{p}{(}\PYG{n}{c} \PYG{o}{+} \PYG{n}{j}\PYG{p}{,} \PYG{l+m+mi}{0}\PYG{p}{),} \PYG{n}{width}\PYG{p}{);}

	\PYG{c+c1}{// Blur is a product of current pixel value and weight of that pixel.}
	\PYG{c+c1}{// Remember that sum of all weights equals to 1,}
	\PYG{c+c1}{// so we are averaging sum of all pixels by their weight.}
	\PYG{k+kt}{float} \PYG{n}{pixel} \PYG{o}{=} \PYG{n}{frame}\PYG{p}{.}\PYG{n}{at}\PYG{o}{\PYGZlt{}}\PYG{n}{uchar}\PYG{o}{\PYGZgt{}}\PYG{p}{(}\PYG{n}{h}\PYG{p}{,}\PYG{n}{w}\PYG{p}{);} \PYG{c+c1}{// (row,col)}
	\PYG{k+kt}{int} \PYG{n}{idx} \PYG{o}{=} \PYG{p}{(}\PYG{n}{i} \PYG{o}{+} \PYG{n}{half}\PYG{p}{)} \PYG{o}{*} \PYG{n}{size}\PYG{p}{.}\PYG{n}{width} \PYG{o}{+} \PYG{p}{(}\PYG{n}{j} \PYG{o}{+} \PYG{n}{half}\PYG{p}{);} \PYG{c+c1}{// width}
	\PYG{k+kt}{float} \PYG{n}{weight} \PYG{o}{=} \PYG{n}{gaussian\PYGZus{}filter}\PYG{p}{[}\PYG{n}{idx}\PYG{p}{];}
	\PYG{n}{blur} \PYG{o}{+=} \PYG{n}{pixel} \PYG{o}{*} \PYG{n}{weight}\PYG{p}{;}
  \PYG{p}{\PYGZcb{}}
\PYG{p}{\PYGZcb{}}
\PYG{n}{destination}\PYG{p}{.}\PYG{n}{at}\PYG{o}{\PYGZlt{}}\PYG{n}{uchar}\PYG{o}{\PYGZgt{}}\PYG{p}{(}\PYG{n}{r}\PYG{p}{,}\PYG{n}{c}\PYG{p}{)} \PYG{o}{=} \PYG{k}{static\PYGZus{}cast}\PYG{o}{\PYGZlt{}}\PYG{k+kt}{unsigned} \PYG{k+kt}{char}\PYG{o}{\PYGZgt{}}\PYG{p}{(}\PYG{n}{blur}\PYG{p}{);}
\end{Verbatim}

\section{Preprocessing Median Blur Code Snippet}
\label{App:preprocmediansnippet}
\begin{Verbatim}[commandchars=\\\{\}]
\PYG{k+kt}{int} \PYG{n}{idx} \PYG{o}{=} \PYG{l+m+mi}{0}\PYG{p}{;}
\PYG{c+c1}{//Average pixel color summing up adjacent pixels.}
\PYG{k}{for} \PYG{p}{(}\PYG{k+kt}{int} \PYG{n}{i} \PYG{o}{=} \PYG{o}{\PYGZhy{}}\PYG{n}{half}\PYG{p}{;} \PYG{n}{i} \PYG{o}{\PYGZlt{}=} \PYG{n}{half}\PYG{p}{;} \PYG{o}{++}\PYG{n}{i}\PYG{p}{)} \PYG{p}{\PYGZob{}}
	\PYG{k}{for} \PYG{p}{(}\PYG{k+kt}{int} \PYG{n}{j} \PYG{o}{=} \PYG{o}{\PYGZhy{}}\PYG{n}{half}\PYG{p}{;} \PYG{n}{j} \PYG{o}{\PYGZlt{}=} \PYG{n}{half}\PYG{p}{;} \PYG{o}{++}\PYG{n}{j}\PYG{p}{)} \PYG{p}{\PYGZob{}}
  		\PYG{c+c1}{// Clamp filter to the image border}
		\PYG{k+kt}{int} \PYG{n}{h} \PYG{o}{=} \PYG{n}{min}\PYG{p}{(}\PYG{n}{max}\PYG{p}{(}\PYG{n}{r} \PYG{o}{+} \PYG{n}{i}\PYG{p}{,} \PYG{l+m+mi}{0}\PYG{p}{),} \PYG{n}{height}\PYG{p}{);}
		\PYG{k+kt}{int} \PYG{n}{w} \PYG{o}{=} \PYG{n}{min}\PYG{p}{(}\PYG{n}{max}\PYG{p}{(}\PYG{n}{c} \PYG{o}{+} \PYG{n}{j}\PYG{p}{,} \PYG{l+m+mi}{0}\PYG{p}{),} \PYG{n}{width}\PYG{p}{);}
		\PYG{n}{window}\PYG{p}{[}\PYG{n}{idx}\PYG{p}{]} \PYG{o}{=} \PYG{n}{frame}\PYG{p}{.}\PYG{n}{at}\PYG{o}{\PYGZlt{}}\PYG{n}{uchar}\PYG{o}{\PYGZgt{}}\PYG{p}{(}\PYG{n}{h}\PYG{p}{,}\PYG{n}{w}\PYG{p}{);} \PYG{c+c1}{// (row,col)}
		\PYG{n}{idx}\PYG{o}{++}\PYG{p}{;}
	\PYG{p}{\PYGZcb{}}
\PYG{p}{\PYGZcb{}}

\PYG{c+c1}{// sort the window to find median}
\PYG{n}{insertionSort}\PYG{p}{(}\PYG{n}{window}\PYG{p}{,} \PYG{n}{window\PYGZus{}len}\PYG{p}{);}

\PYG{c+c1}{// assign the median to centered element of the matrix}
\PYG{n}{destination}\PYG{p}{.}\PYG{n}{at}\PYG{o}{\PYGZlt{}}\PYG{n}{uchar}\PYG{o}{\PYGZgt{}}\PYG{p}{(}\PYG{n}{r}\PYG{p}{,}\PYG{n}{c}\PYG{p}{)} \PYG{o}{=} \PYG{n}{window}\PYG{p}{[}\PYG{n}{idx} \PYG{o}{/} \PYG{l+m+mi}{2}\PYG{p}{];}
\end{Verbatim}

\section{Dual-mode SGM Inner Update Code Snippet}
\label{App:dsgm}
\begin{Verbatim}[commandchars=\\\{\}]
\PYG{n}{cv}\PYG{o}{::}\PYG{n}{Scalar} \PYG{n}{i\PYGZus{}sclr}       \PYG{o}{=} \PYG{n}{next\PYGZus{}frame}\PYG{o}{\PYGZhy{}\PYGZgt{}}\PYG{n}{at}\PYG{o}{\PYGZlt{}}\PYG{n}{uchar}\PYG{o}{\PYGZgt{}}\PYG{p}{(}\PYG{n}{y}\PYG{p}{,}\PYG{n}{x}\PYG{p}{);}
\PYG{n}{cv}\PYG{o}{::}\PYG{n}{Scalar} \PYG{n}{app\PYGZus{}u\PYGZus{}sclr}   \PYG{o}{=} \PYG{n}{app\PYGZus{}u\PYGZus{}mat}\PYG{o}{\PYGZhy{}\PYGZgt{}}\PYG{n}{at}\PYG{o}{\PYGZlt{}}\PYG{n}{uchar}\PYG{o}{\PYGZgt{}}\PYG{p}{(}\PYG{n}{y}\PYG{p}{,}\PYG{n}{x}\PYG{p}{);}
\PYG{n}{cv}\PYG{o}{::}\PYG{n}{Scalar} \PYG{n}{app\PYGZus{}var\PYGZus{}sclr} \PYG{o}{=} \PYG{n}{app\PYGZus{}var\PYGZus{}mat}\PYG{o}{\PYGZhy{}\PYGZgt{}}\PYG{n}{at}\PYG{o}{\PYGZlt{}}\PYG{n}{uchar}\PYG{o}{\PYGZgt{}}\PYG{p}{(}\PYG{n}{y}\PYG{p}{,}\PYG{n}{x}\PYG{p}{);}
\PYG{n}{cv}\PYG{o}{::}\PYG{n}{Scalar} \PYG{n}{can\PYGZus{}u\PYGZus{}sclr}   \PYG{o}{=} \PYG{n}{can\PYGZus{}u\PYGZus{}mat}\PYG{o}{\PYGZhy{}\PYGZgt{}}\PYG{n}{at}\PYG{o}{\PYGZlt{}}\PYG{n}{uchar}\PYG{o}{\PYGZgt{}}\PYG{p}{(}\PYG{n}{y}\PYG{p}{,}\PYG{n}{x}\PYG{p}{);}
\PYG{n}{cv}\PYG{o}{::}\PYG{n}{Scalar} \PYG{n}{can\PYGZus{}var\PYGZus{}sclr} \PYG{o}{=} \PYG{n}{can\PYGZus{}var\PYGZus{}mat}\PYG{o}{\PYGZhy{}\PYGZgt{}}\PYG{n}{at}\PYG{o}{\PYGZlt{}}\PYG{n}{uchar}\PYG{o}{\PYGZgt{}}\PYG{p}{(}\PYG{n}{y}\PYG{p}{,}\PYG{n}{x}\PYG{p}{);}

\PYG{c+c1}{// Get the differences for the candidate and apparent background models}
\PYG{k+kt}{float} \PYG{n}{adiff} \PYG{o}{=} \PYG{n}{i\PYGZus{}sclr}\PYG{p}{.}\PYG{n}{val}\PYG{p}{[}\PYG{l+m+mi}{0}\PYG{p}{]} \PYG{o}{\PYGZhy{}} \PYG{n}{app\PYGZus{}u\PYGZus{}sclr}\PYG{p}{.}\PYG{n}{val}\PYG{p}{[}\PYG{l+m+mi}{0}\PYG{p}{];}
\PYG{k+kt}{float} \PYG{n}{cdiff} \PYG{o}{=} \PYG{n}{i\PYGZus{}sclr}\PYG{p}{.}\PYG{n}{val}\PYG{p}{[}\PYG{l+m+mi}{0}\PYG{p}{]} \PYG{o}{\PYGZhy{}} \PYG{n}{can\PYGZus{}u\PYGZus{}sclr}\PYG{p}{.}\PYG{n}{val}\PYG{p}{[}\PYG{l+m+mi}{0}\PYG{p}{];}

\PYG{k}{if} \PYG{p}{(}\PYG{n}{pow}\PYG{p}{(}\PYG{n}{adiff}\PYG{p}{,} \PYG{l+m+mi}{2}\PYG{p}{)} \PYG{o}{\PYGZlt{}} \PYG{n}{MEAN\PYGZus{}THRESH} \PYG{o}{*} \PYG{n}{max}\PYG{p}{(}\PYG{n}{app\PYGZus{}var\PYGZus{}sclr}\PYG{p}{.}\PYG{n}{val}\PYG{p}{[}\PYG{l+m+mi}{0}\PYG{p}{],} \PYG{l+m+mf}{.1}\PYG{p}{))\PYGZob{}}
    \PYG{n}{i} \PYG{o}{=} \PYG{l+m+mi}{0}\PYG{p}{;}
    \PYG{n}{alpha} \PYG{o}{=} \PYG{l+m+mf}{1.0} \PYG{o}{/} \PYG{p}{(}\PYG{k+kt}{double}\PYG{p}{)}\PYG{n}{app\PYGZus{}ages}\PYG{p}{[}\PYG{n}{x}\PYG{p}{][}\PYG{n}{y}\PYG{p}{];}
    \PYG{n}{app\PYGZus{}u\PYGZus{}sclr}\PYG{p}{.}\PYG{n}{val}\PYG{p}{[}\PYG{n}{i}\PYG{p}{]} \PYG{o}{=} \PYG{p}{(}\PYG{l+m+mf}{1.0}\PYG{o}{\PYGZhy{}}\PYG{n}{alpha}\PYG{p}{)} \PYG{o}{*} \PYG{n}{app\PYGZus{}u\PYGZus{}sclr}\PYG{p}{.}\PYG{n}{val}\PYG{p}{[}\PYG{n}{i}\PYG{p}{]} \PYG{o}{+} \PYG{p}{(}\PYG{n}{alpha}\PYG{p}{)} \PYG{o}{*} \PYG{n}{i\PYGZus{}sclr}\PYG{p}{.}\PYG{n}{val}\PYG{p}{[}\PYG{n}{i}\PYG{p}{];}
    \PYG{n}{V} \PYG{o}{=}  \PYG{n}{pow}\PYG{p}{((}\PYG{n}{app\PYGZus{}u\PYGZus{}sclr}\PYG{p}{.}\PYG{n}{val}\PYG{p}{[}\PYG{n}{i}\PYG{p}{]} \PYG{o}{\PYGZhy{}} \PYG{n}{i\PYGZus{}sclr}\PYG{p}{.}\PYG{n}{val}\PYG{p}{[}\PYG{n}{i}\PYG{p}{]),}\PYG{l+m+mi}{2}\PYG{p}{);}
    \PYG{n}{app\PYGZus{}var\PYGZus{}sclr}\PYG{p}{.}\PYG{n}{val}\PYG{p}{[}\PYG{n}{i}\PYG{p}{]} \PYG{o}{=} \PYG{p}{(}\PYG{l+m+mf}{1.0}\PYG{o}{\PYGZhy{}}\PYG{n}{alpha}\PYG{p}{)} \PYG{o}{*} \PYG{n}{app\PYGZus{}var\PYGZus{}sclr}\PYG{p}{.}\PYG{n}{val}\PYG{p}{[}\PYG{n}{i}\PYG{p}{]} \PYG{o}{+} \PYG{n}{alpha} \PYG{o}{*} \PYG{n}{V}\PYG{p}{;}

    \PYG{c+c1}{//write into matrix}
    \PYG{n}{app\PYGZus{}u\PYGZus{}mat}\PYG{o}{\PYGZhy{}\PYGZgt{}}\PYG{n}{at}\PYG{o}{\PYGZlt{}}\PYG{n}{uchar}\PYG{o}{\PYGZgt{}}\PYG{p}{(}\PYG{n}{y}\PYG{p}{,}\PYG{n}{x}\PYG{p}{)} \PYG{o}{=} \PYG{n}{app\PYGZus{}u\PYGZus{}sclr}\PYG{p}{.}\PYG{n}{val}\PYG{p}{[}\PYG{n}{i}\PYG{p}{];}
    \PYG{n}{app\PYGZus{}var\PYGZus{}mat}\PYG{o}{\PYGZhy{}\PYGZgt{}}\PYG{n}{at}\PYG{o}{\PYGZlt{}}\PYG{n}{uchar}\PYG{o}{\PYGZgt{}}\PYG{p}{(}\PYG{n}{y}\PYG{p}{,}\PYG{n}{x}\PYG{p}{)} \PYG{o}{=} \PYG{n}{app\PYGZus{}var\PYGZus{}sclr}\PYG{p}{.}\PYG{n}{val}\PYG{p}{[}\PYG{n}{i}\PYG{p}{];}

    \PYG{k}{if} \PYG{p}{(}\PYG{n}{app\PYGZus{}ages}\PYG{p}{[}\PYG{n}{x}\PYG{p}{][}\PYG{n}{y}\PYG{p}{]} \PYG{o}{\PYGZlt{}} \PYG{n}{AGE\PYGZus{}THRESH}\PYG{p}{)} \PYG{p}{\PYGZob{}}
        \PYG{n}{app\PYGZus{}ages}\PYG{p}{[}\PYG{n}{x}\PYG{p}{][}\PYG{n}{y}\PYG{p}{]}\PYG{o}{++}\PYG{p}{;}
    \PYG{p}{\PYGZcb{}}

\PYG{p}{\PYGZcb{}} \PYG{k}{else} \PYG{k}{if} \PYG{p}{(}\PYG{n}{pow}\PYG{p}{(}\PYG{n}{cdiff}\PYG{p}{,} \PYG{l+m+mi}{2}\PYG{p}{)} \PYG{o}{\PYGZlt{}} \PYG{n}{MEAN\PYGZus{}THRESH} \PYG{o}{*} \PYG{n}{max}\PYG{p}{(}\PYG{n}{can\PYGZus{}var\PYGZus{}sclr}\PYG{p}{.}\PYG{n}{val}\PYG{p}{[}\PYG{l+m+mi}{0}\PYG{p}{],} \PYG{l+m+mf}{.1}\PYG{p}{))} \PYG{p}{\PYGZob{}}
    \PYG{n}{i} \PYG{o}{=} \PYG{l+m+mi}{0}\PYG{p}{;}
    \PYG{n}{alpha} \PYG{o}{=} \PYG{l+m+mf}{1.0} \PYG{o}{/} \PYG{p}{(}\PYG{k+kt}{double}\PYG{p}{)}\PYG{n}{can\PYGZus{}ages}\PYG{p}{[}\PYG{n}{x}\PYG{p}{][}\PYG{n}{y}\PYG{p}{];}
    \PYG{n}{can\PYGZus{}u\PYGZus{}sclr}\PYG{p}{.}\PYG{n}{val}\PYG{p}{[}\PYG{n}{i}\PYG{p}{]} \PYG{o}{=} \PYG{p}{(}\PYG{l+m+mf}{1.0}\PYG{o}{\PYGZhy{}}\PYG{n}{alpha}\PYG{p}{)} \PYG{o}{*} \PYG{n}{can\PYGZus{}u\PYGZus{}sclr}\PYG{p}{.}\PYG{n}{val}\PYG{p}{[}\PYG{n}{i}\PYG{p}{]} \PYG{o}{+} \PYG{p}{(}\PYG{n}{alpha}\PYG{p}{)} \PYG{o}{*} \PYG{n}{i\PYGZus{}sclr}\PYG{p}{.}\PYG{n}{val}\PYG{p}{[}\PYG{n}{i}\PYG{p}{];}
    \PYG{n}{V} \PYG{o}{=} \PYG{n}{pow}\PYG{p}{((}\PYG{n}{can\PYGZus{}u\PYGZus{}sclr}\PYG{p}{.}\PYG{n}{val}\PYG{p}{[}\PYG{n}{i}\PYG{p}{]} \PYG{o}{\PYGZhy{}} \PYG{n}{i\PYGZus{}sclr}\PYG{p}{.}\PYG{n}{val}\PYG{p}{[}\PYG{n}{i}\PYG{p}{]),}\PYG{l+m+mi}{2}\PYG{p}{);}
    \PYG{n}{can\PYGZus{}var\PYGZus{}sclr}\PYG{p}{.}\PYG{n}{val}\PYG{p}{[}\PYG{n}{i}\PYG{p}{]} \PYG{o}{=} \PYG{p}{(}\PYG{l+m+mf}{1.0}\PYG{o}{\PYGZhy{}}\PYG{n}{alpha}\PYG{p}{)} \PYG{o}{*} \PYG{n}{can\PYGZus{}var\PYGZus{}sclr}\PYG{p}{.}\PYG{n}{val}\PYG{p}{[}\PYG{n}{i}\PYG{p}{]} \PYG{o}{+} \PYG{n}{alpha} \PYG{o}{*} \PYG{n}{V}\PYG{p}{;}

    \PYG{c+c1}{//write into matrix}
    \PYG{n}{can\PYGZus{}u\PYGZus{}mat}\PYG{o}{\PYGZhy{}\PYGZgt{}}\PYG{n}{at}\PYG{o}{\PYGZlt{}}\PYG{n}{uchar}\PYG{o}{\PYGZgt{}}\PYG{p}{(}\PYG{n}{y}\PYG{p}{,}\PYG{n}{x}\PYG{p}{)} \PYG{o}{=} \PYG{n}{can\PYGZus{}u\PYGZus{}sclr}\PYG{p}{.}\PYG{n}{val}\PYG{p}{[}\PYG{n}{i}\PYG{p}{];}
    \PYG{n}{can\PYGZus{}var\PYGZus{}mat}\PYG{o}{\PYGZhy{}\PYGZgt{}}\PYG{n}{at}\PYG{o}{\PYGZlt{}}\PYG{n}{uchar}\PYG{o}{\PYGZgt{}}\PYG{p}{(}\PYG{n}{y}\PYG{p}{,}\PYG{n}{x}\PYG{p}{)} \PYG{o}{=} \PYG{n}{can\PYGZus{}var\PYGZus{}sclr}\PYG{p}{.}\PYG{n}{val}\PYG{p}{[}\PYG{n}{i}\PYG{p}{];}

    \PYG{c+c1}{// Cap ages}
    \PYG{k}{if} \PYG{p}{(}\PYG{n}{can\PYGZus{}ages}\PYG{p}{[}\PYG{n}{x}\PYG{p}{][}\PYG{n}{y}\PYG{p}{]} \PYG{o}{\PYGZlt{}} \PYG{n}{AGE\PYGZus{}THRESH}\PYG{p}{)} \PYG{p}{\PYGZob{}}
        \PYG{n}{can\PYGZus{}ages}\PYG{p}{[}\PYG{n}{x}\PYG{p}{][}\PYG{n}{y}\PYG{p}{]}\PYG{o}{++}\PYG{p}{;}
    \PYG{p}{\PYGZcb{}}

\PYG{p}{\PYGZcb{}} \PYG{k}{else} \PYG{p}{\PYGZob{}}
    \PYG{n}{can\PYGZus{}u\PYGZus{}mat}\PYG{o}{\PYGZhy{}\PYGZgt{}}\PYG{n}{at}\PYG{o}{\PYGZlt{}}\PYG{n}{uchar}\PYG{o}{\PYGZgt{}}\PYG{p}{(}\PYG{n}{y}\PYG{p}{,} \PYG{n}{x}\PYG{p}{)} \PYG{o}{=} \PYG{n}{i\PYGZus{}sclr}\PYG{p}{.}\PYG{n}{val}\PYG{p}{[}\PYG{l+m+mi}{0}\PYG{p}{];}
    \PYG{n}{can\PYGZus{}var\PYGZus{}mat}\PYG{o}{\PYGZhy{}\PYGZgt{}}\PYG{n}{at}\PYG{o}{\PYGZlt{}}\PYG{n}{uchar}\PYG{o}{\PYGZgt{}}\PYG{p}{(}\PYG{n}{y}\PYG{p}{,} \PYG{n}{x}\PYG{p}{)} \PYG{o}{=} \PYG{n}{VAR\PYGZus{}INIT}\PYG{p}{;}
    \PYG{n}{can\PYGZus{}ages}\PYG{p}{[}\PYG{n}{x}\PYG{p}{][}\PYG{n}{y}\PYG{p}{]} \PYG{o}{=} \PYG{l+m+mi}{1}\PYG{p}{;}
\PYG{p}{\PYGZcb{}}

\PYG{k}{if} \PYG{p}{(}\PYG{n}{can\PYGZus{}ages}\PYG{p}{[}\PYG{n}{x}\PYG{p}{][}\PYG{n}{y}\PYG{p}{]} \PYG{o}{\PYGZgt{}} \PYG{n}{app\PYGZus{}ages}\PYG{p}{[}\PYG{n}{x}\PYG{p}{][}\PYG{n}{y}\PYG{p}{])} \PYG{p}{\PYGZob{}}
    \PYG{c+c1}{// Swap the models}
    \PYG{n}{app\PYGZus{}u\PYGZus{}mat}\PYG{o}{\PYGZhy{}\PYGZgt{}}\PYG{n}{at}\PYG{o}{\PYGZlt{}}\PYG{n}{uchar}\PYG{o}{\PYGZgt{}}\PYG{p}{(}\PYG{n}{y}\PYG{p}{,}\PYG{n}{x}\PYG{p}{)} \PYG{o}{=} \PYG{n}{can\PYGZus{}u\PYGZus{}sclr}\PYG{p}{.}\PYG{n}{val}\PYG{p}{[}\PYG{l+m+mi}{0}\PYG{p}{];}
    \PYG{n}{app\PYGZus{}var\PYGZus{}mat}\PYG{o}{\PYGZhy{}\PYGZgt{}}\PYG{n}{at}\PYG{o}{\PYGZlt{}}\PYG{n}{uchar}\PYG{o}{\PYGZgt{}}\PYG{p}{(}\PYG{n}{y}\PYG{p}{,}\PYG{n}{x}\PYG{p}{)} \PYG{o}{=} \PYG{n}{can\PYGZus{}var\PYGZus{}sclr}\PYG{p}{.}\PYG{n}{val}\PYG{p}{[}\PYG{l+m+mi}{0}\PYG{p}{];}
    \PYG{n}{app\PYGZus{}ages}\PYG{p}{[}\PYG{n}{x}\PYG{p}{][}\PYG{n}{y}\PYG{p}{]} \PYG{o}{=} \PYG{n}{can\PYGZus{}ages}\PYG{p}{[}\PYG{n}{x}\PYG{p}{][}\PYG{n}{y}\PYG{p}{];}

    \PYG{c+c1}{//candidateBackgroundModel\PYGZhy{}\PYGZgt{}setPixel(next\PYGZus{}frame, y, x);}
    \PYG{n}{can\PYGZus{}u\PYGZus{}mat}\PYG{o}{\PYGZhy{}\PYGZgt{}}\PYG{n}{at}\PYG{o}{\PYGZlt{}}\PYG{n}{uchar}\PYG{o}{\PYGZgt{}}\PYG{p}{(}\PYG{n}{y}\PYG{p}{,} \PYG{n}{x}\PYG{p}{)} \PYG{o}{=} \PYG{n}{i\PYGZus{}sclr}\PYG{p}{.}\PYG{n}{val}\PYG{p}{[}\PYG{l+m+mi}{0}\PYG{p}{];}
    \PYG{n}{can\PYGZus{}var\PYGZus{}mat}\PYG{o}{\PYGZhy{}\PYGZgt{}}\PYG{n}{at}\PYG{o}{\PYGZlt{}}\PYG{n}{uchar}\PYG{o}{\PYGZgt{}}\PYG{p}{(}\PYG{n}{y}\PYG{p}{,} \PYG{n}{x}\PYG{p}{)} \PYG{o}{=} \PYG{n}{VAR\PYGZus{}INIT}\PYG{p}{;}
    \PYG{n}{can\PYGZus{}ages}\PYG{p}{[}\PYG{n}{x}\PYG{p}{][}\PYG{n}{y}\PYG{p}{]} \PYG{o}{=} \PYG{l+m+mi}{1}\PYG{p}{;}
\PYG{p}{\PYGZcb{}}


\PYG{n}{cv}\PYG{o}{::}\PYG{n}{Scalar} \PYG{n}{app\PYGZus{}diff} \PYG{o}{=} \PYG{n}{app\PYGZus{}u\PYGZus{}mat}\PYG{o}{\PYGZhy{}\PYGZgt{}}\PYG{n}{at}\PYG{o}{\PYGZlt{}}\PYG{n}{uchar}\PYG{o}{\PYGZgt{}}\PYG{p}{(}\PYG{n}{y}\PYG{p}{,}\PYG{n}{x}\PYG{p}{)} \PYG{o}{\PYGZhy{}} \PYG{n}{next\PYGZus{}frame}\PYG{o}{\PYGZhy{}\PYGZgt{}}\PYG{n}{at}\PYG{o}{\PYGZlt{}}\PYG{n}{uchar}\PYG{o}{\PYGZgt{}}\PYG{p}{(}\PYG{n}{y}\PYG{p}{,}\PYG{n}{x}\PYG{p}{);}

\PYG{k}{if} \PYG{p}{(}\PYG{n}{pow}\PYG{p}{(}\PYG{n}{app\PYGZus{}diff}\PYG{p}{.}\PYG{n}{val}\PYG{p}{[}\PYG{l+m+mi}{0}\PYG{p}{],} \PYG{l+m+mi}{2}\PYG{p}{)} \PYG{o}{\PYGZlt{}=} \PYG{n}{THETA\PYGZus{}D}\PYG{o}{*}\PYG{n}{max}\PYG{p}{(}\PYG{l+m+mf}{0.25}\PYG{p}{,} \PYG{n}{i\PYGZus{}sclr}\PYG{p}{.}\PYG{n}{val}\PYG{p}{[}\PYG{l+m+mi}{0}\PYG{p}{]))} \PYG{p}{\PYGZob{}}
    \PYG{c+c1}{//background}
    \PYG{n}{bin\PYGZus{}mat}\PYG{o}{\PYGZhy{}\PYGZgt{}}\PYG{n}{at}\PYG{o}{\PYGZlt{}}\PYG{n}{uchar}\PYG{o}{\PYGZgt{}}\PYG{p}{(}\PYG{n}{y}\PYG{p}{,} \PYG{n}{x}\PYG{p}{)} \PYG{o}{=} \PYG{l+m+mi}{0}\PYG{p}{;}
\PYG{p}{\PYGZcb{}} \PYG{k}{else} \PYG{p}{\PYGZob{}}
    \PYG{c+c1}{//foreground}
    \PYG{n}{bin\PYGZus{}mat}\PYG{o}{\PYGZhy{}\PYGZgt{}}\PYG{n}{at}\PYG{o}{\PYGZlt{}}\PYG{n}{uchar}\PYG{o}{\PYGZgt{}}\PYG{p}{(}\PYG{n}{y}\PYG{p}{,} \PYG{n}{x}\PYG{p}{)} \PYG{o}{=} \PYG{l+m+mi}{255}\PYG{p}{;}
\PYG{p}{\PYGZcb{}}
\end{Verbatim}

\section{Motion Compensation Code Snippets}
\label{App:motion}
\begin{Verbatim}[commandchars=\\\{\}]
\PYG{n}{cv}\PYG{o}{::}\PYG{n}{goodFeaturesToTrack}\PYG{p}{(}\PYG{n}{m\PYGZus{}prevImg}\PYG{p}{,} \PYG{n}{m\PYGZus{}prevPts}\PYG{p}{,} \PYG{n}{maxCorners}\PYG{p}{,}
                            \PYG{n}{qualityLevel}\PYG{p}{,} \PYG{n}{minDistance}\PYG{p}{,} \PYG{n}{mask}\PYG{p}{,} \PYG{n}{blockSize}\PYG{p}{,}
                            \PYG{n}{useHarrisDetector}\PYG{p}{,} \PYG{n}{k}\PYG{p}{);}

\PYG{k}{if}\PYG{p}{(}\PYG{n}{m\PYGZus{}prevPts}\PYG{p}{.}\PYG{n}{size}\PYG{p}{()} \PYG{o}{\PYGZgt{}=} \PYG{l+m+mi}{1}\PYG{p}{)} \PYG{p}{\PYGZob{}}
    \PYG{n}{cv}\PYG{o}{::}\PYG{n}{calcOpticalFlowPyrLK}\PYG{p}{(}\PYG{n}{m\PYGZus{}prevImg}\PYG{p}{,} \PYG{n}{m\PYGZus{}nextImg}\PYG{p}{,} \PYG{n}{m\PYGZus{}prevPts}\PYG{p}{,}
    						\PYG{n}{m\PYGZus{}nextPts}\PYG{p}{,} \PYG{n}{m\PYGZus{}status}\PYG{p}{,} \PYG{n}{m\PYGZus{}error}\PYG{p}{,} \PYG{n}{Size}\PYG{p}{(}\PYG{l+m+mi}{20}\PYG{p}{,}\PYG{l+m+mi}{20}\PYG{p}{),} \PYG{l+m+mi}{5}\PYG{p}{);}

    \PYG{c+c1}{// compute homography using RANSAC}
    \PYG{n}{cv}\PYG{o}{::}\PYG{n}{Mat} \PYG{n}{mask}\PYG{p}{;}
    \PYG{n}{vector} \PYG{o}{\PYGZlt{}}\PYG{n}{Point2f}\PYG{o}{\PYGZgt{}} \PYG{n}{prev\PYGZus{}corner2}\PYG{p}{,} \PYG{n}{cur\PYGZus{}corner2}\PYG{p}{;}
    \PYG{n}{n} \PYG{o}{=} \PYG{n}{next\PYGZus{}frame}\PYG{o}{\PYGZhy{}\PYGZgt{}}\PYG{n}{clone}\PYG{p}{();}

    \PYG{c+c1}{// weed out bad matches}
    \PYG{k}{for}\PYG{p}{(}\PYG{k+kt}{size\PYGZus{}t} \PYG{n}{i}\PYG{o}{=}\PYG{l+m+mi}{0}\PYG{p}{;} \PYG{n}{i} \PYG{o}{\PYGZlt{}} \PYG{n}{m\PYGZus{}status}\PYG{p}{.}\PYG{n}{size}\PYG{p}{();} \PYG{n}{i}\PYG{o}{++}\PYG{p}{)} \PYG{p}{\PYGZob{}}
        \PYG{k}{if}\PYG{p}{(}\PYG{n}{m\PYGZus{}status}\PYG{p}{[}\PYG{n}{i}\PYG{p}{])} \PYG{p}{\PYGZob{}}
            \PYG{n}{prev\PYGZus{}corner2}\PYG{p}{.}\PYG{n}{push\PYGZus{}back}\PYG{p}{(}\PYG{n}{m\PYGZus{}prevPts}\PYG{p}{[}\PYG{n}{i}\PYG{p}{]);}
            \PYG{n}{cur\PYGZus{}corner2}\PYG{p}{.}\PYG{n}{push\PYGZus{}back}\PYG{p}{(}\PYG{n}{m\PYGZus{}nextPts}\PYG{p}{[}\PYG{n}{i}\PYG{p}{]);}
        \PYG{p}{\PYGZcb{}}
    \PYG{p}{\PYGZcb{}}
    \PYG{c+c1}{// apply motion compensation}
    \PYG{n}{cv}\PYG{o}{::}\PYG{n}{Mat} \PYG{n}{H} \PYG{o}{=} \PYG{n}{cv}\PYG{o}{::}\PYG{n}{findHomography}\PYG{p}{(}\PYG{n}{prev\PYGZus{}corner2}\PYG{p}{,}\PYG{n}{cur\PYGZus{}corner2}\PYG{p}{,} \PYG{n}{CV\PYGZus{}RANSAC}\PYG{p}{);}
    \PYG{n}{warpPerspective}\PYG{p}{(}\PYG{n}{m\PYGZus{}nextImg}\PYG{p}{,} \PYG{n}{n}\PYG{p}{,} \PYG{n}{H}\PYG{p}{,} \PYG{n}{m\PYGZus{}prevImg}\PYG{p}{.}\PYG{n}{size}\PYG{p}{(),} \PYG{n}{INTER\PYGZus{}LINEAR} \PYG{o}{|} \PYG{n}{WARP\PYGZus{}INVERSE\PYGZus{}MAP}\PYG{p}{);}
\PYG{p}{\PYGZcb{}}
\end{Verbatim}

\end{appendices}

\end{document}